\documentclass{article}

\PassOptionsToPackage{numbers}{natbib}
\usepackage[eandd, final]{neurips_2026}

\usepackage[utf8]{inputenc} 
\usepackage[T1]{fontenc}    
\usepackage[table]{xcolor}  
\usepackage{hyperref}       
\usepackage{url}            
\usepackage{booktabs}       
\usepackage{amsfonts}       
\usepackage{nicefrac}       
\usepackage{microtype}      
\usepackage{graphicx}
\usepackage{multirow}
\usepackage{tabularx}
\usepackage{adjustbox}
\usepackage{wrapfig}
\usepackage{caption}        
\usepackage{amsmath}
\usepackage{amssymb}
\usepackage{bm}
\usepackage{subfloat}
\usepackage{colortbl}

\newcommand{\fst}[1]{\colorbox{red!30}{#1}}
\newcommand{\snd}[1]{\colorbox{orange!30}{#1}}

\title{PolarScale: A Physics-Grounded Benchmark for Radiometrically Consistent RGB-to-Stokes Estimation}

\author{%
  Beibei Lin$^{1 \dagger}$\enspace
  Tingting Chen$^{1 \dagger}$\enspace
  Xin Zhang$^{1}$\enspace
  Wenhao Zhao$^{1}$\enspace
  Dongjun Li$^{2}$\enspace
  Zifeng Yuan$^{1}$\thanks{Corresponding author: \texttt{zfyuan@nus.edu.sg} \quad $^\dagger$Equal contribution.}\\
  $^{1}$National University of Singapore\quad
  $^{2}$University of Michigan, Ann Arbor
}

\begin{document}

\maketitle

\begin{abstract}
Polarization imaging provides physical cues beyond intensity imaging but typically requires specialized hardware.
Recent methods infer polarization from RGB-like inputs, yet predict only normalized Stokes components or relative descriptors, from which the radiometric scale needed for full Stokes reconstruction has been divided out.
We introduce \textbf{PolarScale}, a benchmark that makes this scale an explicit prediction and evaluation target.
Built on existing trichromatic full-Stokes measurements, PolarScale takes the per-scene normalized total-intensity image $\mathrm{s}_0$ (a scene-referred linear image, not a consumer sRGB photograph) and asks models to predict normalized Stokes components, AoLP/DoLP/DoCP, and a per-scene scale.
Because the scale is divided out of the input, it is not physically identifiable; PolarScale therefore evaluates dataset-conditioned semantic scale estimation against a constant-scale control, together with angular, self-consistency, and physical-bound metrics.
Across seven restoration-based and generative backbones and three prediction strategies, the strongest restoration models estimate the scale with 3.6--4.3\% mean relative error versus 5.7\% for the constant control and violate physical bounds on fewer than 0.25\% of pixels, whereas two generative baselines collapse to a near-zero scale; explicit descriptor supervision improves descriptor accuracy (23.66 vs.\ 18.88 dB PSNR for MAE).
Predicted full-Stokes representations improve diffuse/specular separation, material segmentation, and glare classification, although in diffuse/specular separation the learned scale performs only on par with the constant control.
\end{abstract}

\section{Introduction}
Polarization imaging captures scene properties by measuring the polarization state of light, which is not observable with conventional RGB sensing. This additional polarization information provides cues about physical surface characteristics such as birefringence, surface stress, and micro-scale roughness.
As a result, polarization imaging has enabled progress in challenging vision tasks, including reflection separation~\cite{nayar1997separation, lei2020polarized, pang2022progressive}, material recognition~\cite{wolff1990polarization, mei2022glass, liang2022multimodal, ke2024hydra,ke2025explain}, and shadow removal~\cite{zhou2025polarization}, complementing RGB-based region understanding~\cite{cyclegrpo2026,zhang2024heap} and semantic segmentation~\cite{zhang2025mamba}.
Beyond vision, polarization is also a key degree of freedom in photonic hardware, where the polarization state of lasers can encode binary spins for photonic Ising computing~\cite{gao2024photonic, ke2024hitskt,yuan2025tailored}.

However, acquiring polarization data typically requires dedicated polarization-sensitive hardware such as integrated linear polarizers and retarders~\cite{jeon2024spectral, tu2020divi}, and even controlling polarization at the source requires careful device engineering, such as engineered mesa geometries and integrated metasurfaces in polarization-controlled lasers~\cite{yuan2025mesa, chen2026metasurface}. These systems increase sensor cost, reduce spatial resolution due to multiplexed designs, complicate calibration, and are difficult to integrate into large-scale or mobile vision platforms.
Recovering absolute, radiometrically scaled Stokes vectors is more demanding still, since it additionally requires radiometric calibration of the polarimetric system.
In practical deployment scenarios, such as robotics, autonomous systems, consumer imaging, and wearable sensing~\cite{zhang2018sleep}, these constraints significantly limit the accessibility and scalability of polarization imaging.

\begin{figure*}[t!]
    \centering
    \includegraphics[width=1\linewidth]{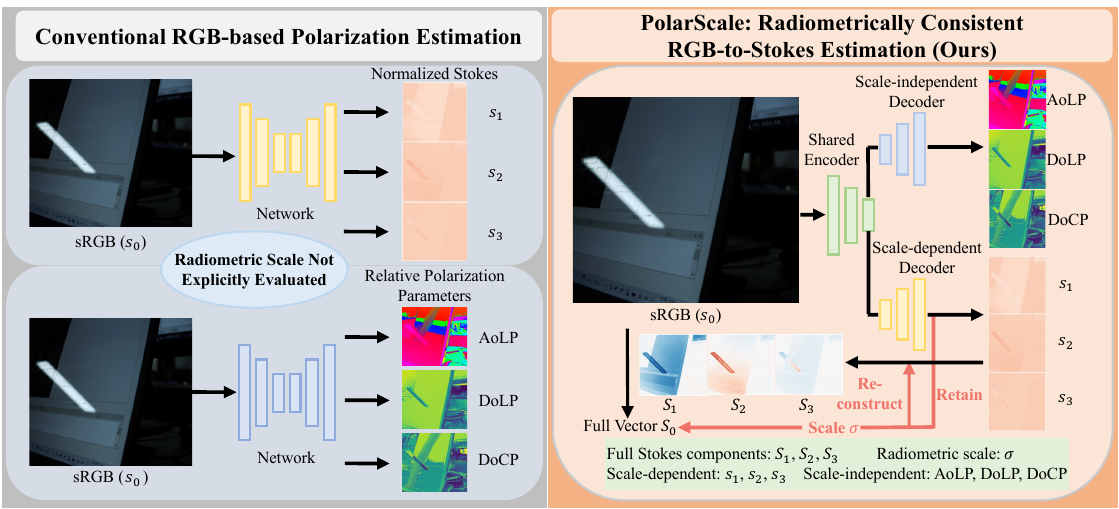}
    \caption{
    Overview of \textbf{PolarScale}.
    Prior RGB-based methods~\cite{linrgb, zhang2025polaranything} predict normalized Stokes components or relative polarimetric descriptors, from which the radiometric scale has been divided out.
    PolarScale additionally predicts the per-scene radiometric scale $\sigma$ and reconstructs the full Stokes components $\mathrm{S}_n = \sigma\,\mathrm{s}_n$, making the scale an explicit evaluation target.
    The input is the normalized total-intensity image $\mathrm{s}_0 = \mathrm{S}_0/\sigma$ (Section~\ref{sec_task}), not a consumer sRGB photograph.
    }
    \label{fig_intro}
    \vspace{-3mm}
\end{figure*}

To overcome these barriers, recent methods infer polarization information directly from RGB-like images using neural networks~\cite{zhang2025polaranything, linrgb}.
However, these approaches predict normalized Stokes components or relative polarimetric descriptors (Figure~\ref{fig_intro}), from which the radiometric scale has been divided out.
Such outputs describe relative polarimetric structure but cannot be turned into absolute Stokes vectors, which radiance-level applications require.
For example, polarization-based diffuse/specular separation attributes the polarized \emph{radiance} to the specular component~\cite{wolff1991constraining, Ke_2025_ICCV_dwim, nayar1997separation}, and glare depends on the amount of polarized light reaching the sensor rather than on the fraction of light that is polarized.
Whether such absolute quantities can be estimated from intensity images has not yet been measured.

In this work, we introduce \textbf{PolarScale}, a benchmark that makes the radiometric scale an explicit prediction and evaluation target.
PolarScale is built on the trichromatic full-Stokes measurements of the Spectral-Polarization dataset~\cite{jeon2024spectral}.
Its input is the normalized total-intensity image $\mathrm{s}_0 = \mathrm{S}_0/\sigma$, a scene-referred linear intensity image rather than a consumer sRGB photograph; we keep the name ``RGB-to-Stokes'' established by~\cite{linrgb} for continuity.
Models predict the normalized Stokes components $\{\mathrm{s}_1, \mathrm{s}_2, \mathrm{s}_3\}$, the descriptors AoLP, DoLP, and DoCP, and the per-scene scale $\sigma$, from which the full Stokes components follow as $\mathrm{S}_n = \sigma\,\mathrm{s}_n$.
Because $\sigma$ is divided out of the input, it cannot be recovered by physical inversion; as in metric monocular depth estimation, it can only be inferred from scene content.
PolarScale therefore measures \emph{dataset-conditioned semantic scale estimation}, and, like diagnostic benchmarks that isolate specific failure modes rather than report only aggregate scores~\cite{chen2025autodiscovery}, we evaluate it accordingly: scale errors are scored against a constant-scale control, and wrap-aware angular, Stokes self-consistency, and physical-bound metrics complement standard image-quality scores.

Benchmarking seven restoration-based and generative backbones under three prediction strategies (direct, joint, and decoupled), we find that the strongest restoration models estimate the scale better than the constant control and nearly satisfy physical bounds, whereas two generative baselines collapse to a near-zero scale, and that full-Stokes PSNR is a poor measure of scale recovery.
Three downstream probes show that predicted full-Stokes representations are useful, while scale-attribution controls show that the contribution of the learned scale itself is mixed, identifying scale accuracy as the open problem PolarScale is designed to track.

Our main contributions are summarized as follows:
\begin{itemize}
\item \textbf{Task reformulation:} PolarScale makes the radiometric scale an explicit prediction and evaluation target of RGB-to-Stokes estimation, and states its input modality and identifiability scope explicitly.

\item \textbf{Physics-aware evaluation protocol:} we score scale estimation against a constant-scale control and add wrap-aware AoLP angular error, Stokes self-consistency, and physical-bound violation metrics. All of them ship in a standalone benchmark package with the preprocessed splits, scale values, derived targets, evaluation scripts, and a datasheet.

\item \textbf{Benchmark analysis and design guidance:} we evaluate seven restoration-based and generative backbones under three prediction strategies, showing that explicit descriptor supervision and decoupled decoding help, and characterizing failure modes such as scale collapse.

\item \textbf{Downstream validation:} we show that predicted full-Stokes representations benefit three downstream tasks, and use scale-attribution controls to separate the contribution of the learned scale from that of the normalized polarization structure.
\end{itemize}

\section{Related work}
\label{sec_related}

\paragraph{\textbf{Polarization datasets}}
Several polarization datasets have been developed to support polarization analysis and related vision tasks.
Early datasets~\cite{ba2020deep, dave2022pandora, liang2022multimodal, qiu2021demos} are mostly small-scale and captured in controlled indoor environments using trichromatic polarization cameras.
Later datasets target specific applications, such as reflection separation~\cite{lei2020polarized, lyu2019refl} and transparent object segmentation~\cite{mei2022glass}.
Fan et al.~\cite{fan2023fullst} introduce a full-Stokes dataset containing both linear and circular polarization components, but it is limited to 67 objects.
Jeon et al.~\cite{jeon2024spectral} provide a large-scale spectro-polarimetric dataset with trichromatic and hyperspectral full-Stokes measurements over more than 2,000 real-world scenes under diverse illumination conditions.
Its full-Stokes measurements retain the radiometric scale, which makes it suitable for constructing PolarScale.

\paragraph{\textbf{RGB-based polarization inference}}
Recent methods infer polarization information directly from RGB inputs~\cite{linrgb, zhang2025polaranything}.
A recent benchmark~\cite{linrgb} establishes a standardized protocol for predicting normalized Stokes components from RGB images, while PolarAnything~\cite{zhang2025polaranything} introduces a diffusion-based framework for polarimetric image synthesis.
Methodologically, existing approaches can be broadly grouped into restoration-based regression and generative image translation.
Restoration-based methods formulate polarization inference as deterministic pixel-wise prediction, using architectures such as Restormer~\cite{zamir2022restormer}, Uformer~\cite{wang2022uformer}, and MAE~\cite{he2022masked}.
Generative methods instead model polarization inference as conditional synthesis, using diffusion-based models such as WDiff~\cite{ozdenizci2023restoring}, DiT~\cite{peebles2023scalable}, RealFill~\cite{tang2024realfill}, and I2ITurbo~\cite{parmar2024one}.
These methods are formulated and evaluated on normalized Stokes components or relative descriptors, so the radiometric scale is never predicted or scored.
Table~\ref{tab_compare} summarizes how PolarScale differs: it is built on the existing measurements of~\cite{jeon2024spectral} rather than on a new capture, and its contribution lies in what is predicted and evaluated.

\begin{table}[t]
  \centering
  \caption{PolarScale compared with related resources. PolarScale re-uses the measurements of~\cite{jeon2024spectral}; it differs in what is predicted and evaluated.}
  \label{tab_compare}
  \small
  \begin{adjustbox}{max width=.98\linewidth}
  \begin{tabular}{lll}
    \toprule
    \textbf{Resource} & \textbf{Prediction target} & \textbf{Radiometric scale evaluated?} \\
    \midrule
    Jeon et al.~\cite{jeon2024spectral} & N/A (physical acquisition, no prediction task) & N/A \\
    Lin et al.~\cite{linrgb} & normalized Stokes $\mathrm{s}_1, \mathrm{s}_2, \mathrm{s}_3$ & No (scale divided out) \\
    PolarAnything~\cite{zhang2025polaranything} & relative descriptors (AoLP, DoLP) & No (scale-invariant by definition) \\
    \textbf{PolarScale (ours)} & $\mathrm{s}_1$--$\mathrm{s}_3$, AoLP/DoLP/DoCP, and $\sigma$ $\rightarrow$ $\mathrm{S}_1$--$\mathrm{S}_3$ & Yes (explicit target, with constant-$\sigma$ control) \\
    \bottomrule
  \end{tabular}
  \end{adjustbox}
  \vspace{-3mm}
\end{table}

\paragraph{\textbf{Polarization-related tasks}}
Polarization provides complementary cues beyond RGB imaging for material, reflection, and geometry analysis.
Because reflected polarization varies with surface properties such as roughness and coating, it has been used for material classification and object recognition~\cite{wolff1990polarization}.
Polarization also separates diffuse and specular reflection~\cite{wolff1991constraining, nayar1997separation}, suppresses specular reflections, and enhances transparent or translucent objects~\cite{riviere2017polarization}.
In geometry estimation, polarization depends on surface orientation and therefore supports surface normal estimation and Shape-from-Polarization reconstruction~\cite{kadambi2015polarized}, complementing intensity-based 3D reconstruction and weather-generalized monocular depth estimation~\cite{yan2025erlora}, whose robustness under degraded capture conditions such as raindrops is itself an active benchmarking topic~\cite{teng2025raindropgs}.
Beyond these tasks, polarization has also been used for reflectance and appearance modeling~\cite{baek2018polarimetric, hwang2022sparse, ha2024polarimetric}, semantic segmentation~\cite{liu2025sharecmp}, sparse sensing~\cite{kurita2023sparse}, color constancy~\cite{ono2022dolp}, and wearable robotics~\cite{yang2018polarization}, and polarization switching in lasers enables drive-efficient GHz-rate binary-state updates~\cite{yuan2026drive} and all-optical annealing~\cite{zhang2025ising}.
Tasks that operate on polarized radiance, such as reflection separation, need absolute rather than normalized Stokes information (Section~\ref{sec_downstream}).

\section{PolarScale benchmark}
\label{sec_benchmark}

\subsection{Preliminaries: Stokes representation}
\label{sec_prelim}
\paragraph{\textbf{Full Stokes parameters}}
We represent the polarization state using the Stokes vector~\cite{linrgb, schaefer2007measuring, yuan2025JLT, sheppard2014jones, yuan2025cleo} (notation in Appendix~\ref{app_notation}), characterized by four parameters $\mathrm{S}_0, \mathrm{S}_1, \mathrm{S}_2, \mathrm{S}_3$. Specifically, $\mathrm{S}_0$ denotes the total light intensity; $\mathrm{S}_1$ corresponds to the intensity difference between $0^\circ$ and $90^\circ$ linear polarization; $\mathrm{S}_2$ represents the difference between $+45^\circ$ and $-45^\circ$ linear polarization; and $\mathrm{S}_3$ indicates the difference between right- and left-handed circular polarization~\cite{mcmaster1954polarization, yuan2025APL}.
A physically realizable Stokes vector satisfies
\begin{equation}
\label{eq:bound}
    \sqrt{\mathrm{S}_1^2 + \mathrm{S}_2^2 + \mathrm{S}_3^2} \le \mathrm{S}_0 .
\end{equation}

\paragraph{\textbf{Interpretable polarization features}}
Commonly used derived representations are the degree of linear polarization (DoLP), the degree of circular polarization (DoCP), and the angle of linear polarization (AoLP)~\cite{tyo2006review}:
\begin{equation}
\label{eq:DoLP/DoCP}
    \text{DoLP} = \frac{\sqrt{\mathrm{S}_1^2 + \mathrm{S}_2^2}}{\mathrm{S}_0}, \quad
    \text{DoCP} = \frac{\mathrm{S}_3}{\mathrm{S}_0},
\end{equation}
\begin{equation}
\label{eq:AoLP}
    \text{AoLP} = \frac{1}{2} \arctan\left( \frac{\mathrm{S}_2}{\mathrm{S}_1} \right).
\end{equation}
DoLP quantifies the proportion of linearly polarized light, whereas DoCP reflects the contribution of circular polarization.
AoLP describes the orientation of the polarization ellipse; it lies in $[-90^\circ, 90^\circ]$ and is periodic with period $180^\circ$, so its errors must be measured with wrap-around (Section~\ref{sec_protocol}).
All three descriptors are ratios of Stokes components and are therefore invariant to the radiometric scale.

\begin{figure*}[t!]
    \centering
    \includegraphics[width=1\linewidth]{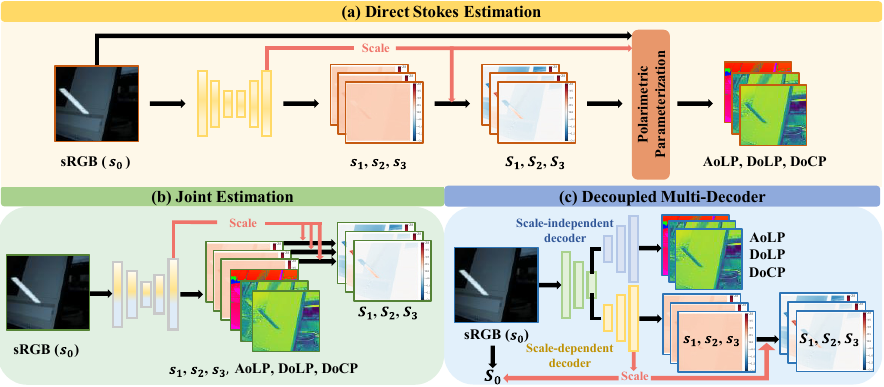}
    \caption{
    Overview of the three prediction strategies evaluated in our benchmark.
    (a) \textbf{Direct Stokes Estimation} predicts $\sigma$ and normalized Stokes components, then derives AoLP, DoLP, and DoCP.
    (b) \textbf{Joint Estimation} predicts all polarimetric parameters with a unified decoder.
    (c) \textbf{Decoupled Multi-decoder} separately estimates scale-dependent components and scale-independent descriptors.
    The input in all three strategies is the normalized total-intensity image $\mathrm{s}_0$.
    }
    \label{fig_architectures}
    \vspace{-5mm}
\end{figure*}

\subsection{Task formulation}
\label{sec_task}

\paragraph{Revisiting polarization inference from RGB images.}
Existing methods learn a mapping $\mathcal{F}_{inv}: \mathbf{I} \to \mathbf{y}_{inv}$ whose output is scale-free: normalized Stokes components $\{\mathrm{s}_1, \mathrm{s}_2, \mathrm{s}_3\}$ in~\cite{linrgb}, or relative descriptors such as AoLP and DoLP in~\cite{zhang2025polaranything}.
Normalized components are obtained from the full components by a normalization operator, either sample-wise min-max normalization~\cite{linrgb} or global scaling by the scene's maximum intensity.
Global scaling preserves the relative radiance ratios within a scene, but in both cases the radiometric scale is removed, so evaluating these outputs cannot assess whether a model recovers absolute Stokes signals.

\paragraph{Input modality.}
For each scene, PolarScale uses global scaling by the per-scene radiometric scale
\begin{equation}
\label{eq:norm}
    \sigma = \max \mathrm{S}_0, \qquad \mathrm{s}_k = \mathrm{S}_k / \sigma, \quad k \in \{0, 1, 2, 3\},
\end{equation}
where the maximum is taken over the scene's $\mathrm{S}_0$ measurement.
The network input is the normalized total-intensity image $\mathrm{s}_0 \in \mathbb{R}^{H \times W \times 3}$ of the trichromatic Stokes arrays of~\cite{jeon2024spectral}.
This is a scene-referred linear intensity image, not a consumer sRGB photograph: it does not pass through the nonlinear tone mapping of a display-referred camera image signal processor (ISP), and it is captured through the polarimetric optics of the source camera.
Following~\cite{linrgb}, we refer to the task as ``RGB-to-Stokes'' estimation; the harder consumer-RGB-to-Stokes problem, which adds ISP nonlinearities and polarization-dependent sensor responses, is outside the scope of the benchmark (Section~\ref{sec_limitations}).

\paragraph{Prediction targets.}
Given $\mathrm{s}_0$, a model predicts
\begin{equation}
    \mathbf{y}_{var} = \{\mathrm{s}_1, \mathrm{s}_2, \mathrm{s}_3, \sigma, \text{AoLP}, \text{DoLP}, \text{DoCP}\},
\end{equation}
that is, a polarimetric representation $\mathbf{P} \in \mathbb{R}^{H \times W \times 19}$ consisting of the normalized Stokes components $\{\mathrm{s}_1, \mathrm{s}_2, \mathrm{s}_3\} \in \mathbb{R}^{H \times W \times 9}$, a scale map $\sigma \in \mathbb{R}^{H \times W \times 1}$ that follows the dense-prediction interface of the baselines, and the descriptors $\{\text{AoLP}, \text{DoLP}, \text{DoCP}\} \in \mathbb{R}^{H \times W \times 9}$.
The full Stokes components are reconstructed as
\begin{equation}
\label{eq:recon}
    \mathrm{S}_n = \sigma \cdot \mathrm{s}_n, \quad n \in \{1, 2, 3\}.
\end{equation}

\paragraph{Identifiability and scope.}
Because $\mathrm{s}_0$ is normalized by the per-scene $\sigma$ (Eq.~\eqref{eq:norm}), $\sigma$ has been divided out of the input and is \emph{not identifiable by physical inversion}: a globally rescaled scene produces the same input.
Since the input is linear rather than display-referred, the ambiguity is a single global factor with no tone-mapping component.
The scale remains learnable only through $p(\sigma \mid \text{scene content})$, because absolute radiance correlates with content such as the illumination type and the indoor/outdoor context; metric monocular depth estimation is ambiguous in the same way and is likewise learned from semantic priors~\cite{bhat2023zoedepth, yin2023metric3d}.
Our scale target draws on this depth-estimation view, in which even the physical disparity of dual-pixel sensors fixes inverse depth only up to an affine transformation~\cite{kurita2025dualpixel}.
Evaluating models on targets that are underdetermined by their inputs is an emerging benchmarking theme in its own right~\cite{chen2026hypospace}.
PolarScale therefore evaluates \emph{dataset-conditioned semantic scale estimation} rather than the recovery of radiometric information contained in the input.
Every scale result must accordingly be read against a constant-scale control that predicts the training prior (Section~\ref{sec_protocol}), and the scale distribution of the source data bounds what the benchmark can measure (Section~\ref{sec_limitations}).

\subsection{Benchmark construction}
\label{sec_benchmark_design}

\paragraph{\textbf{Source data}}
PolarScale is built on the Spectral-Polarization dataset~\cite{jeon2024spectral}, which provides 2,022 trichromatic and 311 hyperspectral full-Stokes images of real-world scenes.
The trichromatic Stokes images have a resolution of $2100 \times 1920$ and are captured in a single shot by a division-of-focal-plane RGB full-Stokes polarimeter with on-sensor quarter-wave plates and linear polarizers~\cite{tu2020divi}, so they include the circular component $\mathrm{S}_3$.
Each scene carries labels for its environment (indoor or outdoor), illumination (clear or cloudy sunlight, white or incandescent light), capture time, and type (object- or scene-oriented).
We use the trichromatic subset, whose full-Stokes measurements retain the radiometric scale needed to define $\sigma$.
Appendix~\ref{app_data} gives further details.

\paragraph{\textbf{Targets and split}}
For each scene, we compute $\sigma$ and the normalized Stokes vector $\{\mathrm{s}_0, \mathrm{s}_1, \mathrm{s}_2, \mathrm{s}_3\}$ by Eq.~\eqref{eq:norm}, and derive AoLP, DoLP, and DoCP by Eqs.~\eqref{eq:DoLP/DoCP} and~\eqref{eq:AoLP}.
The set $\{\sigma, \mathrm{s}_1, \mathrm{s}_2, \mathrm{s}_3, \text{AoLP, DoLP, DoCP}\}$ supervises both scale-dependent and scale-independent quantities.
We adopt the 1,000/200 train/test scene split of Lin et al.~\cite{linrgb}, which keeps our results directly comparable with that protocol; the split is fixed by construction and involves no sampling seed.
The scale distribution is relatively narrow: the training-set median and mean of $\sigma$ are 4.85 and 4.25, and predicting the median for every test scene already gives a 5.7\% mean relative error (Table~\ref{tab_phys}).
Appendix~\ref{app_sigma} summarizes the scale statistics.

\paragraph{\textbf{Prediction strategies}}
No existing method is designed for this setting, so we adapt representative restoration-based and generative architectures from prior RGB-based polarization studies~\cite{linrgb, zhang2025polaranything} and compare three strategies for modeling scale-dependent components $\{\sigma, \mathrm{s}_1, \mathrm{s}_2, \mathrm{s}_3\}$ and scale-independent descriptors $\{\text{AoLP, DoLP, DoCP}\}$ (Figure~\ref{fig_architectures}):

(a) \textbf{Direct Stokes estimation} predicts $\sigma$ and $\{\mathrm{s}_1, \mathrm{s}_2, \mathrm{s}_3\}$ and derives AoLP, DoLP, and DoCP by Eqs.~\eqref{eq:DoLP/DoCP} and~\eqref{eq:AoLP}, which keeps the output compact but depends on the stability of these transformations;
(b) \textbf{joint estimation} predicts all quantities with a unified decoder, so both types are directly supervised;
(c) \textbf{decoupled multi-decoder estimation} uses two parallel decoders for $\{\sigma, \mathrm{s}_1, \mathrm{s}_2, \mathrm{s}_3\}$ and $\{\text{AoLP, DoLP, DoCP}\}$, which encourages specialized representations for physically distinct quantities.

\paragraph{\textbf{Benchmark package}}
We release PolarScale as a standalone package at \url{https://anonymous.4open.science/r/PolarScale}, containing the preprocessed splits and scene lists, per-scene $\sigma$ values, normalized Stokes components and descriptors, evaluation scripts for all metrics of Section~\ref{sec_protocol} (including the constant-$\sigma$ control), baseline code and configurations, and the datasheet of Appendix~\ref{app_datasheet}, so users need not rebuild the benchmark from~\cite{jeon2024spectral}.

\subsection{Evaluation protocol}
\label{sec_protocol}

\paragraph{\textbf{Image-quality metrics}}
Following~\cite{linrgb}, we report PSNR and SSIM for the normalized Stokes components, the descriptors, and the reconstructed full Stokes components, with the PSNR peak fixed to $\mathrm{MAX}_I = 1.0$ for all models and scenes.
Because $\mathrm{s}_n$ is bounded while $\mathrm{S}_n = \sigma\,\mathrm{s}_n$ is not, PSNR values are \emph{not} comparable across these groups, and we do not average them.
LPIPS rescales its inputs and is therefore scale-invariant (identical for $\mathrm{s}_n$ and $\mathrm{S}_n$), so we report it once and do not use it as evidence about scale.

\paragraph{\textbf{Physics-aware metrics}}
Let $\hat{\sigma}$ denote the predicted per-scene scale.
We report:
(i) the \emph{relative scale error} $|\hat{\sigma} - \sigma| / \sigma$ (mean and median over scenes), the \emph{log-scale error} $|\log_{10}\hat{\sigma} - \log_{10}\sigma|$ over scenes with $\hat{\sigma} > 0$, and the number of physically meaningless predictions $\hat{\sigma} \le 0$;
(ii) the \emph{AoLP angular error}, computed with wrap-around at period $180^\circ$ and restricted to pixels with ground-truth DoLP $> 0.1$, since AoLP is undefined for unpolarized light;
(iii) \emph{Stokes self-consistency}: the angular error between the directly predicted AoLP and the AoLP recomputed from the predicted $\mathrm{s}_1, \mathrm{s}_2$, and analogously the mean absolute error for DoLP and DoCP;
(iv) the \emph{physical-bound violation rate}: the fraction of pixels whose predicted Stokes vectors violate Eq.~\eqref{eq:bound}, i.e., $\sqrt{\mathrm{s}_1^2 + \mathrm{s}_2^2 + \mathrm{s}_3^2} > \mathrm{s}_0$.
We measure (iv) on Stokes-derived values because the directly predicted descriptor maps are bounded by their encoding.

\paragraph{\textbf{Constant-scale control}}
To test whether a model learns more than the dataset prior, we pair its predicted normalized Stokes with a constant $\sigma$ equal to the training-set median (4.85) or mean (4.25), and compare the result with the learned $\hat{\sigma}$.
The median control is the reference that any learned scale should beat.

\section{Experiments}
\label{sec_exp}

\subsection{Implementation details}
\label{sec_impl}
\paragraph{\textbf{Baselines}}
We evaluate the restoration-based methods Restormer~\cite{zamir2022restormer}, Uformer~\cite{wang2022uformer}, and MAE~\cite{he2022masked}, and the generative methods WDiff~\cite{ozdenizci2023restoring}, DiT~\cite{peebles2023scalable}, RealFill~\cite{tang2024realfill}, and I2ITurbo~\cite{parmar2024one}, each under the three strategies of Section~\ref{sec_benchmark_design}.

The restoration models, WDiff, and DiT use native multi-channel regression heads whose output matches the strategy (10 channels for (a), 19 for (b) and (c)); for (c) we add a second decoder branch for the descriptors.
For RealFill and I2ITurbo, changing the output dimensionality would disturb the pre-trained weights that are their main strength, so we steer them with task-specific text prompts (e.g., \texttt{"S1"}, \texttt{"Scale"}, \texttt{"AoLP"}) and, for (c), with two independent sets of LoRA parameters.
Their inference uses one pass per component, so every model predicts the same components exactly once.
Appendix~\ref{app_impl} gives the full per-strategy details.

\paragraph{\textbf{Training}}
All baselines were trained to convergence on the same server with eight NVIDIA RTX A5000 GPUs (24 GB each): about one day per restoration model and two days per diffusion-based model.
Per-model hyperparameters are provided in the configuration files of the code release, and Appendix~\ref{app_impl} describes a native-conditioning variant of RealFill and I2ITurbo.
In Table~\ref{tab_quan}, the rows for DiT, I2ITurbo, Uformer, and MAE correspond to their configuration (c) in Table~\ref{tab_ablation}.

\begin{table}[t!]
  \centering
  \caption{
    Image-quality results for normalized Stokes components ($\mathrm{s}_1$--$\mathrm{s}_3$), descriptors, and reconstructed full Stokes components ($\mathrm{S}_1$--$\mathrm{S}_3$).
    PSNR uses $\mathrm{MAX}_I = 1$, so values are not comparable across column groups; $^\dagger$LPIPS is scale-invariant (identical for $\mathrm{s}_n$ and $\mathrm{S}_n$) and reported once.
    Table~\ref{tab_phys} gives the wrap-aware AoLP error.
    \colorbox{red!30}{Best} and \colorbox{orange!30}{second-best} results are highlighted.
    }
  \begin{adjustbox}{max width=.98\linewidth}
    \begin{tabular}{l|ccc|ccc|cc}
    \toprule
    \multicolumn{1}{c|}{\multirow{2}[4]{*}{\textbf{Methods}}} & \multicolumn{3}{c|}{\textbf{$\mathrm{s}_1$, $\mathrm{s}_2$, $\mathrm{s}_3$}} & \multicolumn{3}{c|}{\textbf{AoLP, DoLP, DoCP}} & \multicolumn{2}{c}{\textbf{$\mathrm{S}_1$, $\mathrm{S}_2$, $\mathrm{S}_3$}} \\
\cmidrule{2-9}          & \textbf{PSNR}$\uparrow$ & \textbf{SSIM}$\uparrow$ & \textbf{LPIPS}$^\dagger\downarrow$ & \textbf{PSNR}$\uparrow$ & \textbf{SSIM}$\uparrow$ & \textbf{LPIPS}$\downarrow$ & \textbf{PSNR}$\uparrow$ & \textbf{SSIM}$\uparrow$ \\
    \midrule
    \textbf{WDiff} & 22.39  & 0.6202  & 0.6910  & 12.81  & 0.2800  & 0.7193  & 18.30  & 0.6410  \\
    \textbf{DiT} & 36.32  & 0.9339  & 0.4107  & 18.02  & 0.4547  & 0.5378  & 20.01  & 0.7557  \\
    \textbf{RealFill} & 38.14  & 0.9630  & 0.2078  & 21.23  & 0.5078  & 0.4247  & 16.47  & 0.6109  \\
    \textbf{I2ITurbo} & 38.45  & 0.9735  & 0.3236  & 21.49  & 0.5738  & 0.4891  & 18.47  & 0.6655  \\
    \midrule
    \textbf{Restormer} & 40.22  & 0.9746  & \fst{0.1783}  & 22.54  & 0.6499  & \fst{0.3634}  & 20.21  & \snd{0.7856}  \\
    \textbf{Uformer} & \snd{40.29}  & \snd{0.9755}  & \snd{0.1862}  & \snd{22.80}  & \snd{0.6509}  & 0.3867  & \snd{20.29}  & 0.7853  \\
    \textbf{MAE} & \fst{41.11}  & \fst{0.9790}  & 0.1864  & \fst{23.66}  & \fst{0.6823}  & \snd{0.3725}  & \fst{21.07}  & \fst{0.8136}  \\
    \bottomrule
    \end{tabular}%
  \end{adjustbox}
   \label{tab_quan}%
   \vspace{-3mm}
\end{table}%

\begin{table}[t!]
  \centering
  \caption{
    Physics-aware metrics on the 200 test scenes.
    $\#\hat{\sigma}{\le}0$ counts scenes with a non-positive scale, which are excluded from the log error.
    AoLP errors wrap at $180^\circ$ on pixels with ground-truth DoLP $> 0.1$; self-consistency compares predicted descriptors with those recomputed from the predicted Stokes components; bound violation is the percentage of pixels violating Eq.~\eqref{eq:bound}.
    The constant-$\sigma$ row (training median) is the floor any learned scale should beat.
    }
  \label{tab_phys}
  \small
  \setlength{\tabcolsep}{4pt}
  \begin{adjustbox}{max width=\linewidth}
    \begin{tabular}{l|ccc|cc|c|c}
    \toprule
    \textbf{Method} & \textbf{Rel.\ scale err.\ (\%)} & \textbf{$\log_{10}$} & \textbf{$\#\hat{\sigma}{\le}0$} & \textbf{AoLP} & \textbf{AoLP} & \textbf{DoLP/DoCP} & \textbf{Bound} \\
     & mean / med $\downarrow$ & err.\ $\downarrow$ & $\downarrow$ & err.\ $\downarrow$ & self-cons.\ $\downarrow$ & self-cons.\ $\downarrow$ & viol.\ (\%) $\downarrow$ \\
    \midrule
    \textit{Const.\ $\sigma$ (median)} & \textit{5.7 / 4.5} & -- & -- & -- & -- & -- & -- \\
    \midrule
    \textbf{WDiff} & 92.8 / 93.1 & 1.195 & 11 & 41.0$^\circ$ & 43.7$^\circ$ & 0.442 / 0.613 & 58.1 \\
    \textbf{DiT} & 99.9 / 99.8 & 1.319 & 97 & 38.4$^\circ$ & 36.5$^\circ$ & 0.189 / 0.115 & 5.18 \\
    \textbf{RealFill} & 38.0 / 41.8 & 0.135 & 0 & 32.4$^\circ$ & 44.6$^\circ$ & 0.158 / 0.115 & 3.73 \\
    \textbf{I2ITurbo} & 5.3 / 2.9 & 0.022 & 0 & 32.5$^\circ$ & 52.4$^\circ$ & 0.139 / 0.117 & 3.56 \\
    \midrule
    \textbf{Restormer} & 5.1 / 2.4 & 0.021 & 0 & 31.1$^\circ$ & 24.4$^\circ$ & 0.057 / 0.045 & 0.22 \\
    \textbf{Uformer} & 4.3 / 2.5 & 0.017 & 0 & 30.7$^\circ$ & 30.0$^\circ$ & 0.066 / 0.047 & 0.22 \\
    \textbf{MAE} & \textbf{3.6 / 2.1} & \textbf{0.014} & \textbf{0} & \textbf{26.1}$^\circ$ & \textbf{24.0}$^\circ$ & \textbf{0.054 / 0.045} & \textbf{0.18} \\
    \bottomrule
    \end{tabular}%
  \end{adjustbox}
  \vspace{-3mm}
\end{table}%

\subsection{Main results}
\label{sec_quan}
Table~\ref{tab_quan} reports image-quality metrics and Table~\ref{tab_phys} the physics-aware metrics.
MAE leads on most image-quality columns, consistent with the benefit of pre-trained visual representations for polarization modeling.
Descriptor accuracy remains limited for all models: even with direct supervision, the best AoLP angular error is $26.1^\circ$.

\paragraph{\textbf{Scale estimation against the constant control}}
MAE and Uformer clearly beat the constant-$\sigma$ floor (3.6\% and 4.3\% mean relative error versus 5.7\%), and Restormer and I2ITurbo beat it more narrowly (5.1\% and 5.3\%).
RealFill, DiT, and WDiff fall well below the floor: DiT and WDiff collapse to a near-zero $\hat{\sigma}$, which is non-positive in 97 and 11 of the 200 scenes, while RealFill settles on a large over-estimate.
MAE's estimate is not merely a re-learned constant: the deviation of its $\hat{\sigma}$ from the training prior correlates with the ground-truth deviation ($r = 0.43$), indicating a content-to-scale mapping beyond the dataset average.
However, on scenes whose $\sigma$ departs strongly from the prior, the models regress toward it, as expected for dataset-conditioned scale estimation.

\paragraph{\textbf{Physical consistency}}
The restoration models are nearly physically consistent, with bound violations below 0.25\% of pixels, whereas the generative models violate the bound on 3.6--58\% of pixels.
Self-consistency separates the models in the same way: MAE's directly predicted AoLP deviates by $24.0^\circ$ from the AoLP implied by its own Stokes prediction, versus $52.4^\circ$ for I2ITurbo, whose prompt-conditioned components are predicted in separate passes.
None of these differences is visible in the full-Stokes PSNR of Table~\ref{tab_quan}.

\paragraph{\textbf{Full-Stokes PSNR is not a scale measure}}
The constant-scale control (Appendix~\ref{app_const}, Table~\ref{tab_const}) shows that full-Stokes PSNR barely depends on whether the scale is learned.
Replacing MAE's learned $\hat{\sigma}$ by the training median changes its full-Stokes PSNR from 21.08 to 21.13 dB, and the training mean, whose scale error is about three times larger (16.2\%), yields an even higher PSNR for every model.
With a fixed peak value, PSNR on $\mathrm{S}_n$ is dominated by the normalized structure and rewards shrinking the reconstruction.
The control also exposes the opposite failure: WDiff's collapsed $\hat{\sigma}$ shrinks its reconstructions and inflates its full-Stokes PSNR, which falls from 18.30 to 2.65 dB once an honest constant scale is applied.
We therefore do not interpret the gap between normalized and full-Stokes PSNR in Table~\ref{tab_quan} as evidence about scale recovery, and use the scale metrics of Table~\ref{tab_phys} as the primary scale measures.

\begin{table}[t!]
  \centering
  \caption{Effect of the prediction strategy: (A) direct Stokes estimation, (B) joint estimation, and (C) decoupled multi-decoder estimation, across backbones. Metrics and conventions follow Table~\ref{tab_quan}; PSNR values are not comparable across column groups, and $^\dagger$LPIPS is scale-invariant. Bold marks the best strategy per backbone and column.}
  \begin{adjustbox}{max width=.98\linewidth}
    \begin{tabular}{l|c|ccc|ccc|cc}
    \toprule
    \multicolumn{1}{c|}{\multirow{2}[4]{*}{\textbf{Methods}}} & \multicolumn{1}{c|}{\multirow{2}[4]{*}{\textbf{Type}}} & \multicolumn{3}{c|}{\textbf{$\mathrm{s}_1$, $\mathrm{s}_2$, $\mathrm{s}_3$}} & \multicolumn{3}{c|}{\textbf{AoLP, DoLP, DoCP}} & \multicolumn{2}{c}{\textbf{$\mathrm{S}_1$, $\mathrm{S}_2$, $\mathrm{S}_3$}} \\
\cmidrule{3-10}          &       & \textbf{PSNR}$\uparrow$ & \textbf{SSIM}$\uparrow$ & \textbf{LPIPS}$^\dagger\downarrow$ & \textbf{PSNR}$\uparrow$ & \textbf{SSIM}$\uparrow$ & \textbf{LPIPS}$\downarrow$ & \textbf{PSNR}$\uparrow$ & \textbf{SSIM}$\uparrow$ \\
    \midrule
    \textbf{DiT} & A     & 34.73  & 0.8825  & 0.4548  & 12.79  & \textbf{0.7471}  & \textbf{0.4548}  & \textbf{20.04}  & 0.7471  \\
    \textbf{DiT} & B     & 35.91  & \textbf{0.9484}  & \textbf{0.3468}  & \textbf{18.55}  & 0.4741  & 0.5376  & 20.01  & 0.7446  \\
    \textbf{DiT} & C     & \textbf{36.32}  & 0.9339  & 0.4107  & 18.02  & 0.4547  & 0.5378  & 20.01  & \textbf{0.7557}  \\
    \midrule
    \textbf{I2ITurbo} & A     & \textbf{38.89}  & 0.9733  & \textbf{0.3154}  & 16.48  & 0.3836  & 0.6245  & \textbf{18.93}  & \textbf{0.7019}  \\
    \textbf{I2ITurbo} & B     & 34.98  & 0.9397  & 0.4091  & \textbf{21.77}  & 0.5463  & 0.5399  & 15.17  & 0.5263  \\
    \textbf{I2ITurbo} & C     & 38.45  & \textbf{0.9735}  & 0.3236  & 21.49  & \textbf{0.5738}  & \textbf{0.4891}  & 18.47  & 0.6655  \\
    \midrule
    \textbf{Uformer} & A     & 40.28  & \textbf{0.9756}  & 0.1869  & 19.25  & 0.4920  & 0.4396  & 20.25  & 0.7844  \\
    \textbf{Uformer} & B     & 39.99  & 0.9742  & 0.1954  & 22.73  & 0.6496  & \textbf{0.3866}  & 19.97  & 0.7695  \\
    \textbf{Uformer} & C     & \textbf{40.29}  & 0.9755  & \textbf{0.1862}  & \textbf{22.80}  & \textbf{0.6509}  & 0.3867  & \textbf{20.29}  & \textbf{0.7853}  \\
    \midrule
    \textbf{MAE} & A     & 40.77  & 0.9778  & 0.2456  & 18.88  & 0.5116  & 0.4819  & 20.78  & 0.7940  \\
    \textbf{MAE} & B     & 40.95  & 0.9787  & 0.2251  & 23.54  & 0.6715  & 0.4041  & 20.94  & 0.8078  \\
    \textbf{MAE} & C     & \textbf{41.11}  & \textbf{0.9790}  & \textbf{0.1864}  & \textbf{23.66}  & \textbf{0.6823}  & \textbf{0.3725}  & \textbf{21.07}  & \textbf{0.8136}  \\
    \bottomrule
    \end{tabular}%
    \end{adjustbox}
  \label{tab_ablation}%
  \vspace{-3mm}
\end{table}%

\subsection{Analysis}
\label{sec_dis}
\paragraph{\textbf{Effect of the prediction strategy}}
Table~\ref{tab_ablation} compares the three strategies.
Explicit descriptor supervision is the clearest effect: moving from direct estimation (A) to joint (B) or decoupled (C) estimation improves descriptor PSNR for every backbone (e.g., MAE: 18.88 $\rightarrow$ 23.54/23.66 dB; I2ITurbo: 16.48 $\rightarrow$ 21.77/21.49 dB).
Joint estimation can trade this gain against the scale-dependent outputs, most visibly for I2ITurbo, whose normalized-Stokes PSNR drops from 38.89 to 34.98 dB.
Decoupled estimation avoids this trade-off for the restoration backbones: for MAE it gives the best value in every column, and for Uformer it is the best or within 0.0001 of the best in every column.
We therefore recommend explicit descriptor supervision with decoupled decoding as the default design.

\paragraph{\textbf{Restoration-based versus generative backbones}}
Under our adaptation protocols, the evaluated restoration backbones outperform the evaluated generative ones on normalized-Stokes and descriptor accuracy (Table~\ref{tab_quan}) and on scale estimation and physical consistency (Table~\ref{tab_phys}).
The generative baselines show two distinct failure modes: WDiff and DiT, which use native regression heads, collapse the scale toward zero, and RealFill over-estimates it.
I2ITurbo is the exception on scale (5.3\% error, comparable to Restormer), but it trails on physical consistency.
Retraining RealFill and I2ITurbo with native multi-channel heads that predict all seven outputs in one pass (Appendix~\ref{app_impl}, Table~\ref{tab_native}) proved the weaker adaptation: I2ITurbo's scale error rises from 5.3\% to 8.8\%, and RealFill's native head collapses the scale ($\hat{\sigma} \le 0$ in 21 of 200 scenes; $r = -0.07$ with the true scale); the only gain is self-consistency from single-pass prediction (I2ITurbo: $27.4^\circ$ versus $52.4^\circ$).
We therefore limit this conclusion to the evaluated architectures and adaptation protocols rather than claiming a paradigm-level gap.

\begin{figure*}[t!]
    \centering
    \includegraphics[width=1\linewidth]{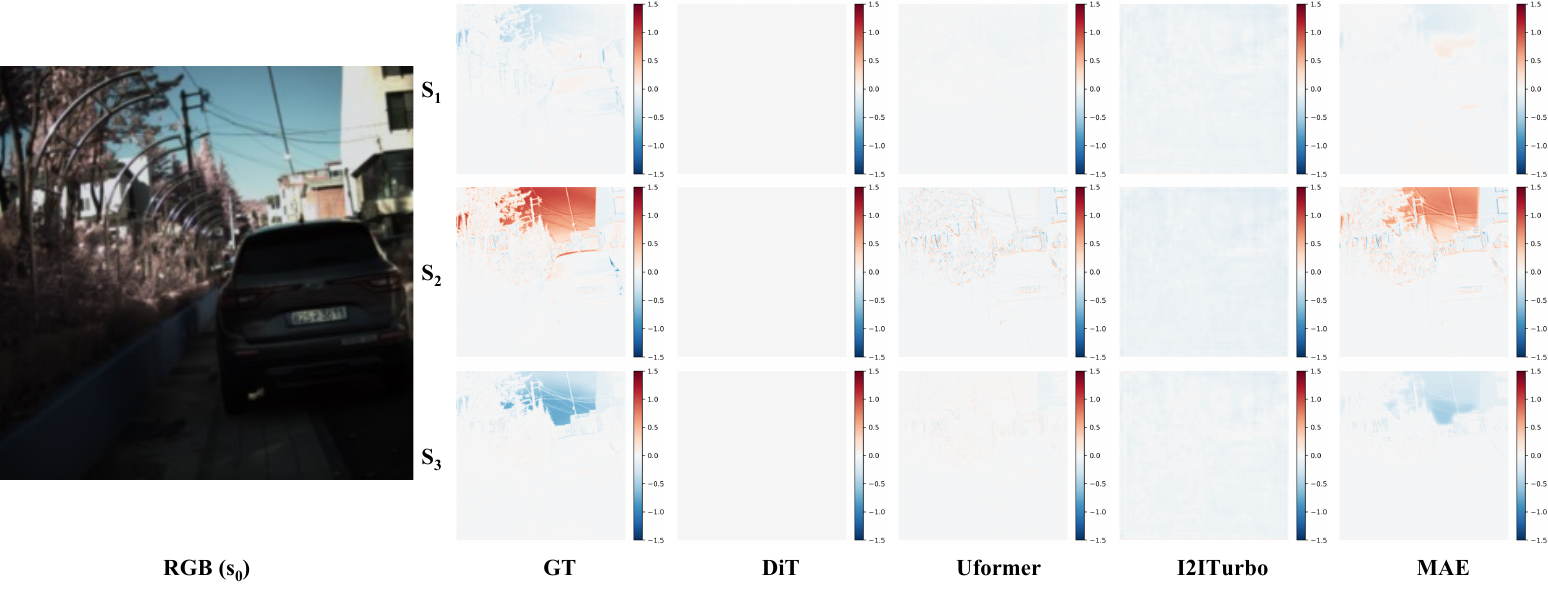}
    \vspace{-6mm}
    \caption{Reconstructed full Stokes components for test scene 1043 (input: normalized $\mathrm{s}_0$). The sky carries strong $\mathrm{S}_2$ and $\mathrm{S}_3$ signals: MAE~\cite{he2022masked} recovers their extent and sign, Uformer~\cite{wang2022uformer} only faint edges, and DiT~\cite{peebles2023scalable} and I2ITurbo~\cite{parmar2024one} predict nearly flat maps.}
    \label{fig_exp1}
    \vspace{-3mm}
\end{figure*}

\subsection{Qualitative evaluation}
\label{sec_qual}
Figure~\ref{fig_exp1} shows a representative result; Appendix~\ref{app_qual} gives further scenes, including descriptor maps (Figures~\ref{fig_dolp_aolp_001} and~\ref{fig_dolp_aolp_002}).
MAE produces the most accurate and spatially consistent predictions and is the only model that recovers the polarized sky in scene 1043, though it predicts the wrong sign of $\mathrm{S}_1$ for the sky in scene 1050.
All models struggle with fine structures and abrupt polarization transitions.

\begin{table}[t!]
  \centering
  \caption{Downstream validation. (a) Diffuse/specular separation, AbsRel $=\mathrm{mean}(|\hat{y} - y| / y)$; the last four rows share the same predicted $\hat{\mathrm{s}}_n$ and differ only in the scale. (b) Material segmentation on MCubeS~\cite{liang2022multimodal}; settings differ only in the polarization input. Glare results are in the text.}
  \label{tab_downstream}
  \small
  \begin{minipage}[t]{0.47\linewidth}
    \centering
    \textbf{(a)} Diffuse/specular separation (AbsRel)\\[2pt]
    \begin{adjustbox}{max width=\linewidth}
    \begin{tabular}{lcc}
      \toprule
      \textbf{Method} & \textbf{Diffuse}$\downarrow$ & \textbf{Specular}$\downarrow$ \\
      \midrule
      RGB-direct regression & 0.190 & 0.508 \\
      $\hat{\mathrm{s}}_n$ + shuffled $\sigma$ & 0.133 & 0.448 \\
      $\hat{\mathrm{s}}_n$ + constant $\sigma$ (median) & 0.118 & 0.448 \\
      $\hat{\mathrm{s}}_n$ + learned $\hat{\sigma}$ (ours) & 0.118 & 0.445 \\
      \midrule
      \textit{$\hat{\mathrm{s}}_n$ + ground-truth $\sigma$} & \textit{0.100} & \textit{0.440} \\
      \bottomrule
    \end{tabular}
    \end{adjustbox}
  \end{minipage}\hfill
  \begin{minipage}[t]{0.49\linewidth}
    \centering
    \textbf{(b)} Material segmentation\\[2pt]
    \begin{adjustbox}{max width=\linewidth}
    \begin{tabular}{lccc}
      \toprule
      \textbf{Polarization input} & \textbf{mIoU}$\uparrow$ & \textbf{Acc$_\text{cls}$}$\uparrow$ & \textbf{fwIoU}$\uparrow$ \\
      \midrule
      Base & 0.322 & 0.421 & 0.600 \\
      + pred.\ DoLP/AoLP & 0.330 & 0.417 & 0.617 \\
      + pred.\ $\hat{\mathrm{s}}_n$ (no scale) & 0.355 & 0.446 & 0.628 \\
      + pred.\ $\hat{\mathrm{s}}_n$ + $\hat{\sigma}$ (ours) & \textbf{0.374} & \textbf{0.468} & \textbf{0.637} \\
      \bottomrule
    \end{tabular}
    \end{adjustbox}
  \end{minipage}
  \vspace{-3mm}
\end{table}

\subsection{Downstream validation}
\label{sec_downstream}
\paragraph{\textbf{Tasks}}
We run three probes of whether predicted absolute Stokes information is useful beyond reconstruction quality (Table~\ref{tab_downstream}; protocols in Appendix~\ref{app_downstream}).
(i) \emph{Diffuse/specular separation.} Following the classical polarization-based separation~\cite{wolff1991constraining, nayar1997separation}, the polarized part of the reflected radiance is attributed to the specular component and the unpolarized remainder to the diffuse component, both in absolute radiance.
A baseline regresses the six absolute channels (diffuse and specular RGB) directly, whereas we compose them analytically from the predicted $\hat{\mathrm{s}}_n$ and a scale; holding $\hat{\mathrm{s}}_n$ fixed, we vary the scale source (shuffled across scenes, constant training median, learned $\hat{\sigma}$, or ground truth as an upper bound) to attribute its effect.
(ii) \emph{Material segmentation} on MCubeS~\cite{liang2022multimodal}, whose 20 classes are annotated independently of polarization: a base input without polarization is augmented with different predicted polarization representations.
(iii) \emph{Glare-level classification.} Glare is an inherently absolute quantity: it depends on the amount of polarized light reaching the sensor, not on the fraction of light that is polarized. Each test scene receives a glare level from the absolute polarized radiance of its ground-truth Stokes vectors, which is predicted either from image appearance alone or from the predicted absolute Stokes.

\paragraph{\textbf{Results}}
The predicted full-Stokes representation improves all three tasks over the tested RGB or relative-polarization baselines: diffuse/specular AbsRel drops from 0.190/0.508 to 0.118/0.445, mIoU rises from 0.330 with predicted DoLP/AoLP to 0.374, and glare accuracy over the 200 test scenes rises from 48.5\% (Wilson 95\% interval~\cite{wilson1927probable} [41.7, 55.4]) with image appearance to 63.0\% ([56.1, 69.4]) with predicted absolute Stokes.
The contribution of the learned scale itself is mixed.
In diffuse/specular separation, the learned $\hat{\sigma}$ performs on par with the constant control (0.118/0.445 versus 0.118/0.448) and thus does not consistently beat the dataset prior, whereas the ground-truth scale gives a clear improvement (0.100/0.440) and a shuffled scale degrades the diffuse estimate (0.133): an accurate absolute scale matters, but current estimates are not yet accurate enough to exploit it.
In material segmentation, adding $\hat{\sigma}$ to the predicted normalized Stokes raises mIoU from 0.355 to 0.374, but without constant- or shuffled-scale variants this does not uniquely isolate the per-scene scale; the glare probe has no scale control.
Our claim is therefore that predicted full-Stokes representations benefit the evaluated tasks, with the absolute scale as one contributing component whose accuracy remains the main open problem.

\section{Limitations and broader impacts}
\label{sec_limitations}
\paragraph{\textbf{Scale is not physically identifiable}}
Because $\sigma$ is divided out of the input, PolarScale measures dataset-conditioned semantic scale estimation; predicted scales and reconstructed $\mathrm{S}_n$ are not calibrated radiometry, and models regress toward the training prior on scenes whose scale departs strongly from it.

\paragraph{\textbf{Single source dataset and narrow scale distribution}}
All data come from one dataset~\cite{jeon2024spectral} captured with one polarimetric camera, and its scale distribution is narrow: a constant predictor already reaches 5.7\% mean relative error, leaving a small margin for learned models (3.6\% at best).
Scale results may not transfer to other sensors, exposures, or scene distributions, motivating domain generalization and test-time adaptation~\cite{zhang2023adaptive}.

\paragraph{\textbf{Input modality and scene conditions}}
The input is a normalized linear image captured through polarimetric optics; consumer sRGB images add ISP nonlinearities (tone mapping, white balance, clipping) and polarization-dependent sensor responses.
No dataset pairs consumer RGB with absolute-scale Stokes ground truth, so this setting cannot yet be evaluated quantitatively.
Low light, strong specular highlights, and adverse weather~\cite{chen2024dual, lin2024nightrain, lin2025nighthaze}, for which dedicated benchmarks exist in other vision tasks~\cite{yang2025erf, ke2026view2space,du2026weatherreasonseg}, are not covered.

\paragraph{\textbf{Scope of the conclusions}}
The restoration-versus-generative comparison holds only for the evaluated architectures and adaptation protocols.
All results come from a single training run per configuration on a fixed split; we report Wilson intervals for the glare probe.
The downstream probes are small-scale, two lack full scale controls, and the scale-attribution result is mixed.
All models still fail on fine structures and abrupt polarization transitions (Figures~\ref{fig_exp1} and~\ref{fig_dolp_aolp_001}; AoLP errors above $26^\circ$), which physics-informed constraints may address.

\paragraph{\textbf{Broader impacts}}
\label{sec_impact}
PolarScale may benefit polarization-aware perception in robotics, material analysis, remote sensing, 3D reconstruction~\cite{lin2026glowgs}, image completion~\cite{lin2026geocomplete} and computational imaging by reducing reliance on specialized hardware, with medical image segmentation~\cite{qian2024msnsegnet} and whole-slide classification~\cite{zhang2022gigapixel} as further applications to investigate, and lightweight estimators could eventually run on emerging photonic accelerators for convolutional networks~\cite{liu2025lithium}.
Its main risk is over-reliance on inferred polarization in safety-critical settings such as driver assistance: predicted scales are dataset-conditioned estimates, not measurements, and should be validated under the target sensor, lighting, and domain before high-stakes use.
The benchmark is not designed for person identification or surveillance.

\section{Conclusion}
We introduced \textbf{PolarScale}, a benchmark that makes the radiometric scale an explicit target of RGB-to-Stokes estimation and, because the scale is divided out of the input, evaluates dataset-conditioned scale estimation against a constant-scale control together with angular, self-consistency, and physical-bound metrics.
The strongest evaluated restoration models beat the dataset prior and nearly satisfy physical bounds, two generative baselines collapse the scale, and explicit descriptor supervision with decoupled decoding gives the most balanced predictions.
Predicted full-Stokes representations benefit three downstream tasks, while the learned scale is not yet accurate enough to consistently beat the prior; PolarScale makes this gap measurable for future physically grounded RGB-to-Stokes models.

{
\small
\bibliographystyle{unsrt}
\bibliography{main}

\begin{thebibliography}{10}

\bibitem{nayar1997separation}
Shree~K. Nayar, Xi-Sheng Fang, and Terrance Boult.
\newblock Separation of reflection components using color and polarization.
\newblock {\em International Journal of Computer Vision}, 21(3):163--186, 1997.

\bibitem{lei2020polarized}
Chenyang Lei, Xuhua Huang, Mengdi Zhang, Qiong Yan, Wenxiu Sun, and Qifeng
  Chen.
\newblock Polarized reflection removal with perfect alignment in the wild.
\newblock In {\em Proceedings of the IEEE/CVF Conference on Computer Vision and
  Pattern Recognition}, 2020.

\bibitem{pang2022progressive}
Youxin Pang, Mengke Yuan, Qiang Fu, Peiran Ren, and Dong-Ming Yan.
\newblock Progressive polarization based reflection removal via realistic
  training data generation.
\newblock {\em Pattern Recognition}, 124:108497, 2022.

\bibitem{wolff1990polarization}
Lawrence~B. Wolff.
\newblock Polarization-based material classification from specular reflection.
\newblock {\em IEEE Transactions on Pattern Analysis and Machine Intelligence},
  12(11):1059--1071, 1990.

\bibitem{mei2022glass}
Haiyang Mei, Bo~Dong, Wen Dong, Jiaxi Yang, Seung-Hwan Baek, Felix Heide,
  Pieter Peers, Xiaopeng Wei, and Xin Yang.
\newblock Glass segmentation using intensity and spectral polarization cues.
\newblock In {\em Proceedings of the IEEE/CVF Conference on Computer Vision and
  Pattern Recognition}, 2022.

\bibitem{liang2022multimodal}
Yupeng Liang, Ryosuke Wakaki, Shohei Nobuhara, and Ko~Nishino.
\newblock Multimodal material segmentation.
\newblock In {\em Proceedings of the IEEE/CVF Conference on Computer Vision and
  Pattern Recognition}, 2022.

\bibitem{ke2024hydra}
Fucai Ke, Zhixi Cai, Simindokht Jahangard, Weiqing Wang, Pari~Delir Haghighi,
  and Hamid Rezatofighi.
\newblock {HYDRA}: A hyper agent for dynamic compositional visual reasoning.
\newblock In {\em European Conference on Computer Vision}, pages 132--149.
  Springer, 2024.

\bibitem{ke2025explain}
Fucai Ke, Joy Hsu, Zhixi Cai, Zixian Ma, Xin Zheng, Xindi Wu, Sukai Huang,
  Weiqing Wang, Pari~Delir Haghighi, Gholamreza Haffari, et~al.
\newblock Explain before you answer: A survey on compositional visual
  reasoning.
\newblock {\em arXiv preprint arXiv:2508.17298}, 2025.

\bibitem{zhou2025polarization}
Chu Zhou, Chao Xu, and Boxin Shi.
\newblock Polarization guided mask-free shadow removal.
\newblock In {\em Proceedings of the AAAI Conference on Artificial
  Intelligence}, volume~39, pages 10716--10724, 2025.

\bibitem{cyclegrpo2026}
Xin Zhang, Haochen Wang, Yikang Zhou, Zhuochen Wang, Xiangtai Li, and Robby~T.
  Tan.
\newblock Actor as its own critic: Unifying region understanding and
  localization via {CycleGRPO}.
\newblock In {\em European Conference on Computer Vision (ECCV)}, pages
  586--604. Springer, 2026.

\bibitem{zhang2024heap}
Xin Zhang, Jinheng Xie, Yuan Yuan, Michael~Bi Mi, and Robby~T. Tan.
\newblock {HEAP}: Unsupervised object discovery and localization with
  contrastive grouping.
\newblock In {\em Proceedings of the AAAI Conference on Artificial
  Intelligence}, volume~38, pages 7323--7331, 2024.

\bibitem{zhang2025mamba}
Xin Zhang and Robby~T. Tan.
\newblock Mamba as a bridge: Where vision foundation models meet vision
  language models for domain-generalized semantic segmentation.
\newblock In {\em Proceedings of the IEEE/CVF Conference on Computer Vision and
  Pattern Recognition}, 2025.

\bibitem{gao2024photonic}
Yuan Gao, Guanyu Chen, Luo Qi, Wujie Fu, Zifeng Yuan, and Aaron~J. Danner.
\newblock Photonic {Ising} machines for combinatorial optimization problems.
\newblock {\em Applied Physics Reviews}, 11(4):041307, 2024.

\bibitem{ke2024hitskt}
Fucai Ke, Weiqing Wang, Weicong Tan, Lan Du, Yuan Jin, Yujin Huang, and Hongzhi
  Yin.
\newblock {HiTSKT}: A hierarchical transformer model for session-aware
  knowledge tracing.
\newblock {\em Knowledge-Based Systems}, 284:111300, 2024.

\bibitem{yuan2025tailored}
Zifeng Yuan, Tingting Chen, Dewen Zhang, Yuan Gao, Wenkai Shan, Beibei Lin, and
  Aaron Danner.
\newblock Tailored polarization-switchable {VCSEL} arrays for photonic {Ising}
  computing.
\newblock {\em Applied Physics Letters}, 127(22):221102, 2025.

\bibitem{jeon2024spectral}
Yujin Jeon, Eunsue Choi, Youngchan Kim, Yunseong Moon, Khalid Omer, Felix
  Heide, and Seung-Hwan Baek.
\newblock Spectral and polarization vision: Spectro-polarimetric real-world
  dataset.
\newblock In {\em Proceedings of the IEEE/CVF Conference on Computer Vision and
  Pattern Recognition}, 2024.

\bibitem{tu2020divi}
Xingzhou Tu, Scott McEldowney, Yang Zou, Matthew Smith, Christopher Guido, Neal
  Brock, Sawyer Miller, Linan Jiang, and Stanley Pau.
\newblock Division of focal plane red--green--blue full-{S}tokes imaging
  polarimeter.
\newblock {\em Applied Optics}, 59(22):G33--G40, 2020.

\bibitem{yuan2025mesa}
Zifeng Yuan, Wenkai Shan, Tingting Chen, Beibei Lin, and Aaron Danner.
\newblock Mesa orientation engineering for polarization locking in {VCSEL}s.
\newblock In {\em 2025 IEEE Photonics Conference (IPC)}, pages 1--2, 2025.

\bibitem{chen2026metasurface}
Wenjie Chen, Zifeng Yuan, Hong-Lin Lin, Luo Qi, Jiaru Chu, Aaron Danner, and
  Yuhang Chen.
\newblock Metasurface-integrated {VCSEL} designed for polarization control in
  optical {Ising} machines.
\newblock {\em arXiv preprint arXiv:2609.21331}, 2026.

\bibitem{zhang2018sleep}
Xin Zhang, Weixuan Kou, Eric I-Chao Chang, He~Gao, Yubo Fan, and Yan Xu.
\newblock Sleep stage classification based on multi-level feature learning and
  recurrent neural networks via wearable device.
\newblock {\em Computers in Biology and Medicine}, 103:71--81, 2018.

\bibitem{zhang2025polaranything}
Kailong Zhang, Youwei Lyu, Heng Guo, Si~Li, Zhanyu Ma, and Boxin Shi.
\newblock {PolarAnything}: Diffusion-based polarimetric image synthesis.
\newblock In {\em Proceedings of the IEEE/CVF International Conference on
  Computer Vision}, 2025.

\bibitem{linrgb}
Beibei Lin, Zifeng Yuan, and Tingting Chen.
\newblock {RGB}-to-polarization estimation: A new task and benchmark study.
\newblock In {\em Advances in Neural Information Processing Systems (Datasets
  and Benchmarks Track)}, volume~38, 2025.

\bibitem{wolff1991constraining}
Lawrence~B. Wolff and Terrance~E. Boult.
\newblock Constraining object features using a polarization reflectance model.
\newblock {\em IEEE Transactions on Pattern Analysis and Machine Intelligence},
  13(7):635--657, 1991.

\bibitem{Ke_2025_ICCV_dwim}
Fucai Ke, Vijay Kumar~B G, Xingjian Leng, Zhixi Cai, Zaid Khan, Weiqing Wang,
  Pari~Delir Haghighi, Hamid Rezatofighi, and Manmohan Chandraker.
\newblock {DWIM}: Towards tool-aware visual reasoning via discrepancy-aware
  workflow generation \& instruct-masking tuning.
\newblock In {\em Proceedings of the IEEE/CVF International Conference on
  Computer Vision (ICCV)}, 2025.

\bibitem{chen2025autodiscovery}
Tingting Chen, Beibei Lin, Srinivas Anumasa, Vedant Shah, Zifeng Yuan, Qiran
  Zou, Anirudh Goyal, and Dianbo Liu.
\newblock {Auto-Discovery-Bench}: Diagnosing structured state tracking in
  oracle-guided discovery.
\newblock {\em arXiv preprint arXiv:2502.15224}, 2025.

\bibitem{ba2020deep}
Yunhao Ba, Alex Gilbert, Franklin Wang, Jinfa Yang, Rui Chen, Yiqin Wang, Lei
  Yan, Boxin Shi, and Achuta Kadambi.
\newblock Deep shape from polarization.
\newblock In {\em European Conference on Computer Vision}, pages 554--571.
  Springer, 2020.

\bibitem{dave2022pandora}
Akshat Dave, Yongyi Zhao, and Ashok Veeraraghavan.
\newblock {PANDORA}: Polarization-aided neural decomposition of radiance.
\newblock In {\em European Conference on Computer Vision}, pages 538--556.
  Springer, 2022.

\bibitem{qiu2021demos}
Simeng Qiu, Qiang Fu, Congli Wang, and Wolfgang Heidrich.
\newblock Linear polarization demosaicking for monochrome and colour
  polarization focal plane arrays.
\newblock {\em Computer Graphics Forum}, 40(6):77--89, 2021.

\bibitem{lyu2019refl}
Youwei Lyu, Zhaopeng Cui, Si~Li, Marc Pollefeys, and Boxin Shi.
\newblock Reflection separation using a pair of unpolarized and polarized
  images.
\newblock In H.~Wallach, H.~Larochelle, A.~Beygelzimer, F.~d\textquotesingle
  Alch\'{e}-Buc, E.~Fox, and R.~Garnett, editors, {\em Advances in Neural
  Information Processing Systems}, volume~32. Curran Associates, Inc., 2019.

\bibitem{fan2023fullst}
Axin Fan, Tingfa Xu, Geer Teng, Xi~Wang, Yuhan Zhang, Chang Xu, Xin Xu, and
  Jianan Li.
\newblock Full-{S}tokes polarization multispectral images of various
  stereoscopic objects.
\newblock {\em Scientific Data}, 10(1):328, 2023.

\bibitem{zamir2022restormer}
Syed~Waqas Zamir, Aditya Arora, Salman Khan, Munawar Hayat, Fahad~Shahbaz Khan,
  and Ming-Hsuan Yang.
\newblock Restormer: Efficient transformer for high-resolution image
  restoration.
\newblock In {\em Proceedings of the IEEE/CVF Conference on Computer Vision and
  Pattern Recognition}, 2022.

\bibitem{wang2022uformer}
Zhendong Wang, Xiaodong Cun, Jianmin Bao, Wengang Zhou, Jianzhuang Liu, and
  Houqiang Li.
\newblock Uformer: A general {U}-shaped transformer for image restoration.
\newblock In {\em Proceedings of the IEEE/CVF Conference on Computer Vision and
  Pattern Recognition}, 2022.

\bibitem{he2022masked}
Kaiming He, Xinlei Chen, Saining Xie, Yanghao Li, Piotr Doll{\'a}r, and Ross
  Girshick.
\newblock Masked autoencoders are scalable vision learners.
\newblock In {\em Proceedings of the IEEE/CVF Conference on Computer Vision and
  Pattern Recognition}, 2022.

\bibitem{ozdenizci2023restoring}
Ozan {\"O}zdenizci and Robert Legenstein.
\newblock Restoring vision in adverse weather conditions with patch-based
  denoising diffusion models.
\newblock {\em IEEE Transactions on Pattern Analysis and Machine Intelligence},
  45(8):10346--10357, 2023.

\bibitem{peebles2023scalable}
William Peebles and Saining Xie.
\newblock Scalable diffusion models with transformers.
\newblock In {\em Proceedings of the IEEE/CVF International Conference on
  Computer Vision}, 2023.

\bibitem{tang2024realfill}
Luming Tang, Nataniel Ruiz, Qinghao Chu, Yuanzhen Li, Aleksander Holynski,
  David~E. Jacobs, Bharath Hariharan, Yael Pritch, Neal Wadhwa, Kfir Aberman,
  et~al.
\newblock {RealFill}: Reference-driven generation for authentic image
  completion.
\newblock {\em ACM Transactions on Graphics (TOG)}, 43(4):1--12, 2024.

\bibitem{parmar2024one}
Gaurav Parmar, Taesung Park, Srinivasa Narasimhan, and Jun-Yan Zhu.
\newblock One-step image translation with text-to-image models.
\newblock {\em arXiv preprint arXiv:2403.12036}, 2024.

\bibitem{riviere2017polarization}
J{\'e}r{\'e}my Riviere, Ilya Reshetouski, Luka Filipi, and Abhijeet Ghosh.
\newblock Polarization imaging reflectometry in the wild.
\newblock {\em ACM Transactions on Graphics}, 36(6):1--14, 2017.

\bibitem{kadambi2015polarized}
Achuta Kadambi, Vage Taamazyan, Boxin Shi, and Ramesh Raskar.
\newblock Polarized {3D}: High-quality depth sensing with polarization cues.
\newblock In {\em Proceedings of the IEEE International Conference on Computer
  Vision}, 2015.

\bibitem{yan2025erlora}
Weilong Yan, Xin Zhang, and Robby~T. Tan.
\newblock {ER-LoRA}: Effective-rank guided adaptation for weather-generalized
  depth estimation.
\newblock {\em arXiv preprint arXiv:2509.00665}, 2025.

\bibitem{teng2025raindropgs}
Zhiqiang Teng, Tingting Chen, Beibei Lin, Zifeng Yuan, Xuanyi Li, Xuanyu Zhang,
  and Shunli Zhang.
\newblock {RaindropGS}: A benchmark for {3D} {Gaussian} splatting under
  raindrop conditions.
\newblock {\em arXiv preprint arXiv:2510.17719}, 2025.

\bibitem{baek2018polarimetric}
Seung-Hwan Baek, Daniel~S. Jeon, Xin Tong, and Min~H. Kim.
\newblock Simultaneous acquisition of polarimetric {SVBRDF} and normals.
\newblock {\em ACM Transactions on Graphics}, 37(6):1--15, 2018.

\bibitem{hwang2022sparse}
Inseung Hwang, Daniel~S. Jeon, Adolfo Mu{\~n}oz, Diego Gutierrez, Xin Tong, and
  Min~H. Kim.
\newblock Sparse ellipsometry: portable acquisition of polarimetric {SVBRDF}
  and shape with unstructured flash photography.
\newblock {\em ACM Transactions on Graphics}, 41(4):1--14, 2022.

\bibitem{ha2024polarimetric}
Hyunho Ha, Inseung Hwang, Nestor Monzon, Jaemin Cho, Donggun Kim, Seung-Hwan
  Baek, Adolfo Mu{\~n}oz, Diego Gutierrez, and Min~H. Kim.
\newblock Polarimetric {BSSRDF} acquisition of dynamic faces.
\newblock {\em ACM Transactions on Graphics}, 43(6):1--11, 2024.

\bibitem{liu2025sharecmp}
Zhuoyan Liu, Bo~Wang, Lizhi Wang, Chenyu Mao, and Ye~Li.
\newblock {ShareCMP}: Polarization-aware {RGB-P} semantic segmentation.
\newblock {\em IEEE Transactions on Circuits and Systems for Video Technology},
  35(10):10316--10329, 2025.

\bibitem{kurita2023sparse}
Teppei Kurita, Yuhi Kondo, Legong Sun, and Yusuke Moriuchi.
\newblock Simultaneous acquisition of high quality {RGB} image and polarization
  information using a sparse polarization sensor.
\newblock In {\em Proceedings of the IEEE/CVF Winter Conference on Applications
  of Computer Vision (WACV)}, 2023.

\bibitem{ono2022dolp}
Taishi Ono, Yuhi Kondo, Legong Sun, Teppei Kurita, and Yusuke Moriuchi.
\newblock Degree-of-linear-polarization-based color constancy.
\newblock In {\em Proceedings of the IEEE/CVF Conference on Computer Vision and
  Pattern Recognition}, 2022.

\bibitem{yang2018polarization}
Kailun Yang, Luis~M. Bergasa, Eduardo Romera, Xiao Huang, and Kaiwei Wang.
\newblock Predicting polarization beyond semantics for wearable robotics.
\newblock In {\em Proc. IEEE-RAS Int. Conf. Humanoid Robots (Humanoids)}, pages
  96--103. IEEE, 2018.

\bibitem{yuan2026drive}
Zifeng Yuan, Beibei Lin, Yutong Liu, Wujie Fu, Yuan Gao, Tingting Chen, and
  Aaron Danner.
\newblock Drive-efficient {GHz}-rate binary-state updates enabled by {VCSEL}
  polarization switching.
\newblock {\em Laser \& Photonics Reviews}, page e71890, 2026.

\bibitem{zhang2025ising}
Dewen Zhang, Zifeng Yuan, Thanh~Xuan Hoang, Wujie Fu, Ching~Eng Png, Soon~Thor
  Lim, and Aaron Danner.
\newblock All-optical scalable and programmable {VCSEL}-based {Ising} annealer
  with parallel feedback.
\newblock {\em Optics Express}, 33(11):22119--22131, 2025.

\bibitem{schaefer2007measuring}
Beth Schaefer, Edward Collett, Robert Smyth, Daniel Barrett, and Beth Fraher.
\newblock Measuring the {S}tokes polarization parameters.
\newblock {\em American Journal of Physics}, 75(2):163--168, 2007.

\bibitem{yuan2025JLT}
Zifeng Yuan, Dewen Zhang, Yuan Gao, Luo Qi, Wujie Fu, and Aaron Danner.
\newblock Large-scale fabrication and analysis of polarization behavior in
  {VCSEL}s with tailored apertures.
\newblock {\em Journal of Lightwave Technology}, 43(14):6819--6827, 2025.

\bibitem{sheppard2014jones}
C.~J.~R. Sheppard.
\newblock {J}ones and {S}tokes parameters for polarization in three dimensions.
\newblock {\em Physical Review A}, 90(2):023809, 2014.

\bibitem{yuan2025cleo}
Zifeng Yuan, Dewen Zhang, Hong-Lin Lin, and Aaron Danner.
\newblock Engineering polarization switching in {VCSEL}s with custom aperture
  shapes.
\newblock In {\em CLEO: Applications and Technology}, page JPS200\_47. Optica
  Publishing Group, 2025.

\bibitem{mcmaster1954polarization}
W.~H. McMaster.
\newblock Polarization and the {S}tokes parameters.
\newblock {\em American Journal of Physics}, 22(6):351--362, 1954.

\bibitem{yuan2025APL}
Zifeng Yuan, Dewen Zhang, Lei Shi, Yutong Liu, and Aaron Danner.
\newblock Enhanced polarization locking in {VCSEL}s.
\newblock {\em Applied Physics Letters}, 126(15):151101, 2025.

\bibitem{tyo2006review}
J.~Scott Tyo, Dennis~L. Goldstein, David~B. Chenault, and Joseph~A. Shaw.
\newblock Review of passive imaging polarimetry for remote sensing
  applications.
\newblock {\em Applied Optics}, 45(22):5453--5469, 2006.

\bibitem{bhat2023zoedepth}
Shariq~Farooq Bhat, Reiner Birkl, Diana Wofk, Peter Wonka, and Matthias
  M{\"u}ller.
\newblock {ZoeDepth}: Zero-shot transfer by combining relative and metric
  depth.
\newblock {\em arXiv preprint arXiv:2302.12288}, 2023.

\bibitem{yin2023metric3d}
Wei Yin, Chi Zhang, Hao Chen, Zhipeng Cai, Gang Yu, Kaixuan Wang, Xiaozhi Chen,
  and Chunhua Shen.
\newblock {Metric3D}: Towards zero-shot metric {3D} prediction from a single
  image.
\newblock In {\em Proceedings of the IEEE/CVF International Conference on
  Computer Vision}, 2023.

\bibitem{kurita2025dualpixel}
Teppei Kurita, Yuhi Kondo, Legong Sun, Takayuki Sasaki, Sho Nitta, Yasuhiro
  Hashimoto, Yoshinori Muramatsu, and Yusuke Moriuchi.
\newblock Revisiting disparity from dual-pixel images: Physics-informed
  lightweight depth estimation.
\newblock In {\em Proceedings of the IEEE/CVF Winter Conference on Applications
  of Computer Vision (WACV)}, 2025.

\bibitem{chen2026hypospace}
Tingting Chen, Beibei Lin, Zifeng Yuan, Qiran Zou, Hongyu He, Anirudh Goyal,
  Yew-Soon Ong, and Dianbo Liu.
\newblock {HypoSpace}: A diagnostic benchmark for set-valued hypothesis
  generation under underdetermination and sublinear coverage bounds.
\newblock In {\em Proceedings of the 43rd International Conference on Machine
  Learning}, volume 306 of {\em Proceedings of Machine Learning Research},
  pages 15451--15465. PMLR, 2026.

\bibitem{wilson1927probable}
Edwin~B. Wilson.
\newblock Probable inference, the law of succession, and statistical inference.
\newblock {\em Journal of the American Statistical Association},
  22(158):209--212, 1927.

\bibitem{zhang2023adaptive}
Xin Zhang and Ying-Cong Chen.
\newblock Adaptive domain generalization via online disagreement minimization.
\newblock {\em IEEE Transactions on Image Processing}, 32:4247--4258, 2023.

\bibitem{chen2024dual}
Tingting Chen, Beibei Lin, Yeying Jin, Wending Yan, Wei Ye, Yuan Yuan, and
  Robby~T. Tan.
\newblock {Dual-Rain}: Video rain removal using assertive and gentle teachers.
\newblock In {\em European Conference on Computer Vision}, pages 127--143.
  Springer, 2024.

\bibitem{lin2024nightrain}
Beibei Lin, Yeying Jin, Wending Yan, Wei Ye, Yuan Yuan, Shunli Zhang, and
  Robby~T. Tan.
\newblock {NightRain}: Nighttime video deraining via adaptive-rain-removal and
  adaptive-correction.
\newblock In {\em Proceedings of the AAAI Conference on Artificial
  Intelligence}, volume~38, pages 3378--3385, 2024.

\bibitem{lin2025nighthaze}
Beibei Lin, Yeying Jin, Wending Yan, Wei Ye, Yuan Yuan, and Robby~T. Tan.
\newblock {NightHaze}: Nighttime image dehazing via self-prior learning.
\newblock In {\em Proceedings of the AAAI Conference on Artificial
  Intelligence}, volume~39, pages 5209--5217, 2025.

\bibitem{yang2025erf}
Xin Yang, Xin Zhang, and Xinchao Wang.
\newblock {ERF}: A benchmark dataset for robust semantic segmentation under
  extreme rainfall conditions.
\newblock In {\em Proceedings of the AAAI Conference on Artificial
  Intelligence}, volume~39, pages 9301--9309, 2025.

\bibitem{ke2026view2space}
Fucai Ke, Zhixi Cai, Boying Li, Long Chen, Beibei Lin, Weiqing Wang, Pari~Delir
  Haghighi, Gholamreza Haffari, and Hamid Rezatofighi.
\newblock {VIEW2SPACE}: Studying multi-view visual reasoning from sparse
  observations.
\newblock In {\em European Conference on Computer Vision}, pages 520--538.
  Springer, 2026.

\bibitem{du2026weatherreasonseg}
Wanjun Du, Zifeng Yuan, Tingting Chen, Fucai Ke, Beibei Lin, and Shunli Zhang.
\newblock {WeatherReasonSeg}: A benchmark for weather-aware reasoning
  segmentation in visual language models.
\newblock In {\em European Conference on Computer Vision (ECCV)}, pages 1--19.
  Springer, 2026.

\bibitem{lin2026glowgs}
Beibei Lin, Xiao Cao, Jingyuan Guo, and Robby~T. Tan.
\newblock {GlowGS}: Generative semantic feature learning for {3D} {Gaussian}
  splatting in nighttime glow scenes.
\newblock {\em arXiv preprint arXiv:2605.23602}, 2026.

\bibitem{lin2026geocomplete}
Beibei Lin, Tingting Chen, and Robby Tan.
\newblock {GeoComplete}: Geometry-aware diffusion for reference-driven image
  completion.
\newblock In {\em Advances in Neural Information Processing Systems},
  volume~38, 2025.

\bibitem{qian2024msnsegnet}
Ziniu Qian, Zihua Wang, Xin Zhang, Bingzheng Wei, Maode Lai, Jianzhong Shou,
  Yubo Fan, and Yan Xu.
\newblock {MSNSegNet}: Attention-based multi-shape nuclei instance segmentation
  in histopathology images.
\newblock {\em Medical \& Biological Engineering \& Computing},
  62(6):1821--1836, 2024.

\bibitem{zhang2022gigapixel}
Jingwei Zhang, Xin Zhang, Ke~Ma, Rajarsi Gupta, Joel Saltz, Maria Vakalopoulou,
  and Dimitris Samaras.
\newblock Gigapixel whole-slide images classification using locally supervised
  learning.
\newblock In {\em Medical Image Computing and Computer Assisted Intervention --
  MICCAI 2022}, pages 192--201. Springer Nature Switzerland, 2022.

\bibitem{liu2025lithium}
Yutong Liu, Pragati Aashna, Zifeng Yuan, Wujie Fu, Mingxiang Yang, Yunjie Yan,
  Luo Qi, Mingshan Zhao, and Aaron Danner.
\newblock A lithium niobate cascaded {MMI}-based arbitrary matrix calculator
  for convolutional neural networks.
\newblock {\em Optics \& Laser Technology}, 192:113740, 2025.

\bibitem{gebru2021datasheets}
Timnit Gebru, Jamie Morgenstern, Briana Vecchione, Jennifer~Wortman Vaughan,
  Hanna Wallach, Hal Daum{\'e}~III, and Kate Crawford.
\newblock Datasheets for datasets.
\newblock {\em Communications of the ACM}, 64(12):86--92, 2021.

\end{thebibliography}
}


\clearpage
\appendix

\section{Notation}
\label{app_notation}

\begin{table}[h]
  \centering
  \caption{Notation used in this paper.}
  \label{tab_notation}
  \small
  \begin{tabular}{ll}
    \toprule
    \textbf{Symbol} & \textbf{Meaning} \\
    \midrule
    $\mathrm{S}_0, \mathrm{S}_1, \mathrm{S}_2, \mathrm{S}_3$ & Full (absolute) Stokes components, trichromatic ($H \times W \times 3$ each) \\
    $\sigma$ & Per-scene radiometric scale, $\sigma = \max \mathrm{S}_0$ \\
    $\hat{\sigma}$ & Predicted per-scene scale \\
    $\mathrm{s}_k = \mathrm{S}_k/\sigma$ & Normalized Stokes components, $k \in \{0,1,2,3\}$ \\
    $\mathrm{s}_0$ & Network input: normalized total-intensity image \\
    $\hat{\mathrm{s}}_n$ & Predicted normalized Stokes components, $n \in \{1,2,3\}$ \\
    AoLP, DoLP, DoCP & Angle / degree of linear and degree of circular polarization (Eqs.~\eqref{eq:DoLP/DoCP}, \eqref{eq:AoLP}) \\
    $\mathbf{P}$ & Network output tensor ($H \times W \times 10$ for type (a), $H \times W \times 19$ for types (b), (c)) \\
    $\mathrm{MAX}_I$ & PSNR peak value, fixed to $1.0$ \\
    Constant $\sigma$ & Training-set median (4.85) or mean (4.25) of $\sigma$ \\
    \bottomrule
  \end{tabular}
\end{table}

\section{Source dataset and benchmark details}
\label{app_data}

\subsection{Source dataset}
The Spectral-Polarization dataset~\cite{jeon2024spectral} contains 2,022 trichromatic Stokes images ($2100 \times 1920 \times 3 \times 4$) and 311 hyperspectral Stokes images ($612 \times 512 \times 21 \times 4$, 450--650\,nm in 10\,nm steps).
The trichromatic images are captured in a single shot by a division-of-focal-plane RGB full-Stokes polarimeter with on-sensor quarter-wave plates and linear polarizers~\cite{tu2020divi}; the hyperspectral images are captured by scanning a liquid-crystal tunable filter and rotating a quarter-wave plate.
Every image is labeled with its environment (indoor/outdoor), illumination (clear or cloudy sunlight, white or incandescent light), capture time, and scene type (object- or scene-oriented).
PolarScale uses only the trichromatic subset.

\subsection{Split and package contents}
The 1,000/200 split follows the scene split of Lin et al.~\cite{linrgb}; it is fixed by construction, and no additional sampling seed is involved.
The benchmark package lists the scene IDs of both splits, the per-scene $\sigma$ values, and all derived targets, so that every number in this paper can be reproduced from the released files.

\subsection{Scale distribution}
\label{app_sigma}
The training-set median and mean of $\sigma$ are 4.85 and 4.25; using them as constant predictors gives mean/median relative errors of 5.7\%/4.5\% and 16.2\%/16.0\% on the test split.
Since the mean lies below the median, the distribution has a tail toward small scales.

\section{Implementation details}
\label{app_impl}

\paragraph{\textbf{Training budget}}
All baselines were trained to convergence on one server with eight NVIDIA RTX A5000 GPUs (24 GB each): about one day for each restoration model and two days for each diffusion-based model.
Per-model hyperparameters are provided in the configuration files of the code release.

\paragraph{\textbf{Output heads per strategy}}
\textit{Type (a), direct Stokes estimation.}
The output tensor is $\mathbf{P} \in \mathbb{R}^{H \times W \times 10}$, concatenating $\{\mathrm{s}_1, \mathrm{s}_2, \mathrm{s}_3\} \in \mathbb{R}^{H \times W \times 9}$ and the scale map $\sigma \in \mathbb{R}^{H \times W \times 1}$.
For the restoration-based methods, WDiff, and DiT, we set the dimensionality of the final output layer to 10.
For RealFill and I2ITurbo, we use task-specific text prompts (\texttt{"S1"}, \texttt{"S2"}, \texttt{"S3"}, \texttt{"Scale"}) with image-polarization pairs to guide training and steer inference.
\textit{Type (b), joint estimation.}
The output tensor is $\mathbf{P} \in \mathbb{R}^{H \times W \times 19}$, adding AoLP, DoLP, and DoCP; RealFill and I2ITurbo receive the additional prompts \texttt{"AoLP"}, \texttt{"DoLP"}, and \texttt{"DoCP"}.
\textit{Type (c), decoupled multi-decoder estimation.}
For the restoration-based methods, WDiff, and DiT, we add a second decoder branch: the original one estimates scale-dependent parameters and the new one scale-independent parameters.
For RealFill and I2ITurbo, we use two independent sets of LoRA parameters instead of an additional decoder, which avoids the parameter redundancy and overhead of extra frozen decoder layers.

\paragraph{\textbf{Generative adaptation}}
WDiff and DiT are not prompt-adapted: they use native multi-channel regression heads.
RealFill and I2ITurbo use text prompts, because changing their output dimensionality would disturb the large-scale pre-trained weights that are their main strength.
Their inference is multi-pass, one pass per component, and is matched at the task level: every model predicts the same components exactly once.

\paragraph{\textbf{Native-conditioning variant}}
To test whether prompt-based adaptation disadvantaged the pre-trained generative baselines, we also trained them with native conditioning.
The text encoder is frozen and receives an empty prompt, and all seven outputs are predicted jointly in a single forward pass through widened layers (RealFill: \texttt{conv\_in} $9 \rightarrow 33$ and \texttt{conv\_out} $4 \rightarrow 28$ channels; I2ITurbo: \texttt{conv\_out} $4 \rightarrow 28$ channels), initialized by replicating the pre-trained weights.
The training recipe is otherwise unchanged (same LoRA ranks, optimizer, learning rate, batch size, and resolution).
Table~\ref{tab_native} shows that native conditioning is the weaker adaptation.

\begin{table}[h]
  \centering
  \caption{Prompt-based versus native conditioning for the pre-trained generative baselines.}
  \label{tab_native}
  \begin{adjustbox}{max width=.98\linewidth}
  \begin{tabular}{lcccc}
    \toprule
     & \textbf{I2ITurbo (prompt)} & \textbf{I2ITurbo (native)} & \textbf{RealFill (prompt)} & \textbf{RealFill (native)} \\
    \midrule
    Rel.\ scale err.\ (mean/med) $\downarrow$ & \textbf{5.3\% / 2.9\%} & 8.8\% / 7.4\% & 38.0\% / 41.8\% & 65.9\% / 69.7\% \\
    Normalized PSNR / SSIM $\uparrow$ & \textbf{38.46 / 0.974} & 35.35 / 0.961 & 38.14 / 0.963 & 29.66 / 0.901 \\
    $\#(\hat{\sigma} \le 0)$ $\downarrow$ & 0/200 & 0/200 & 0/200 & 21/200 \\
    AoLP self-consistency $\downarrow$ & 52.4$^\circ$ & \textbf{27.4}$^\circ$ & 44.6$^\circ$ & 36.3$^\circ$ \\
    \bottomrule
  \end{tabular}
  \end{adjustbox}
\end{table}

\section{Constant-scale control on full-Stokes reconstruction}
\label{app_const}

Table~\ref{tab_const} combines each model's predicted normalized Stokes with either its learned $\hat{\sigma}$ or a constant scale (training median 4.85 or mean 4.25).

\begin{table}[h]
  \centering
  \caption{Full-Stokes PSNR/SSIM with the learned scale and with constant scales. For the well-behaved models, the constant scale matches or exceeds the learned scale in full-Stokes PSNR, so this metric does not measure scale recovery. WDiff and DiT expose the opposite failure: their near-zero $\hat{\sigma}$ inflates the PSNR obtained with the learned scale.}
  \label{tab_const}
  \begin{adjustbox}{max width=.98\linewidth}
  \begin{tabular}{lcccc}
    \toprule
    \textbf{Method} & \textbf{Rel.\ scale err.\ (mean/med)}$\downarrow$ & \textbf{Learned $\hat{\sigma}$} & \textbf{Const.\ $\sigma = 4.85$} & \textbf{Const.\ $\sigma = 4.25$} \\
    \midrule
    WDiff & 92.8\% / 93.1\% & 18.30 / 0.641 & 2.65 / 0.087 & 3.74 / 0.105 \\
    DiT & 99.9\% / 99.8\% & 20.01 / 0.746 & 16.44 / 0.607 & 17.06 / 0.633 \\
    RealFill & 38.0\% / 41.8\% & 16.47 / 0.611 & 18.25 / 0.684 & 18.78 / 0.707 \\
    I2ITurbo & 5.3\% / 2.9\% & 18.47 / 0.666 & 18.52 / 0.669 & 18.89 / 0.691 \\
    Restormer & 5.1\% / 2.4\% & 20.21 / 0.786 & 20.25 / 0.786 & 20.50 / 0.788 \\
    Uformer & 4.3\% / 2.5\% & 20.29 / 0.785 & 20.31 / 0.786 & 20.50 / 0.787 \\
    MAE & \textbf{3.6\% / 2.1\%} & 21.08 / 0.814 & 21.13 / 0.815 & 21.27 / 0.816 \\
    \bottomrule
  \end{tabular}
  \end{adjustbox}
\end{table}

\section{Downstream protocols}
\label{app_downstream}

\paragraph{\textbf{Diffuse/specular separation}}
Reference diffuse and specular components are computed from the measured full Stokes vectors of the 200 test scenes by the same polarization-based separation, in absolute radiance.
Our setting composes both components analytically from $\hat{\mathrm{s}}_n$ and a scale; the RGB-direct baseline regresses the six absolute channels directly.
For the shuffled-scale control, the learned scales are permuted across test scenes.

\paragraph{\textbf{Material segmentation}}
All settings share the same network, training data, schedule, and initialization, and differ only in the polarization representation supplied.

\paragraph{\textbf{Glare-level classification}}
The 200 scenes are the benchmark's held-out test split, and their glare levels are computed from the ground-truth polarization of those same scenes; no component of this probe is re-fit, so there is no variation across training seeds or split resampling.
The intervals reported in Section~\ref{sec_downstream} are Wilson 95\% score intervals for a binomial proportion over the 200 scenes~\cite{wilson1927probable}.

\section{Datasheet}
\label{app_datasheet}
We document the PolarScale benchmark package following Datasheets for Datasets~\cite{gebru2021datasheets}.

\paragraph{Motivation}
PolarScale was created to evaluate whether models can estimate radiometrically scaled Stokes vectors from intensity images, a question that existing benchmarks, which evaluate only normalized or relative polarization, cannot answer. It was created by the authors of this paper.

\paragraph{Composition}
Each of the 1,200 instances (1,000 train, 200 test) is a real-world scene from the trichromatic subset of~\cite{jeon2024spectral}.
For every scene, the package provides the input $\mathrm{s}_0$, the scale $\sigma$, the normalized Stokes components $\mathrm{s}_1$--$\mathrm{s}_3$, the descriptors AoLP, DoLP, and DoCP, and the scene labels inherited from~\cite{jeon2024spectral}.
The data contain no annotations of people. Outdoor scenes may incidentally show vehicles or bystanders, as in the source dataset.

\paragraph{Collection process}
No new data were captured. All instances are derived from~\cite{jeon2024spectral} by the deterministic preprocessing of Section~\ref{sec_benchmark_design}.

\paragraph{Preprocessing}
$\sigma = \max \mathrm{S}_0$ per scene; $\mathrm{s}_k = \mathrm{S}_k/\sigma$; AoLP, DoLP, and DoCP by Eqs.~\eqref{eq:DoLP/DoCP} and~\eqref{eq:AoLP}. The preprocessing scripts are part of the package.

\paragraph{Uses}
Intended for benchmarking RGB-to-Stokes estimation with an explicit scale target. Not intended for calibrated radiometry, safety-critical decisions without validation on the target sensor, or the identification of people.

\paragraph{Distribution}
The package is distributed through the code repository given in Appendix~\ref{app_ethics}. The derived data remain subject to the license and terms of use of the source dataset~\cite{jeon2024spectral}.

\paragraph{Maintenance}
The package is maintained by the authors; questions and error reports can be sent to the corresponding author (\texttt{zfyuan@nus.edu.sg}).

\section{Additional qualitative results}
\label{app_qual}

\begin{figure*}[h]
    \centering
    \includegraphics[width=1\linewidth]{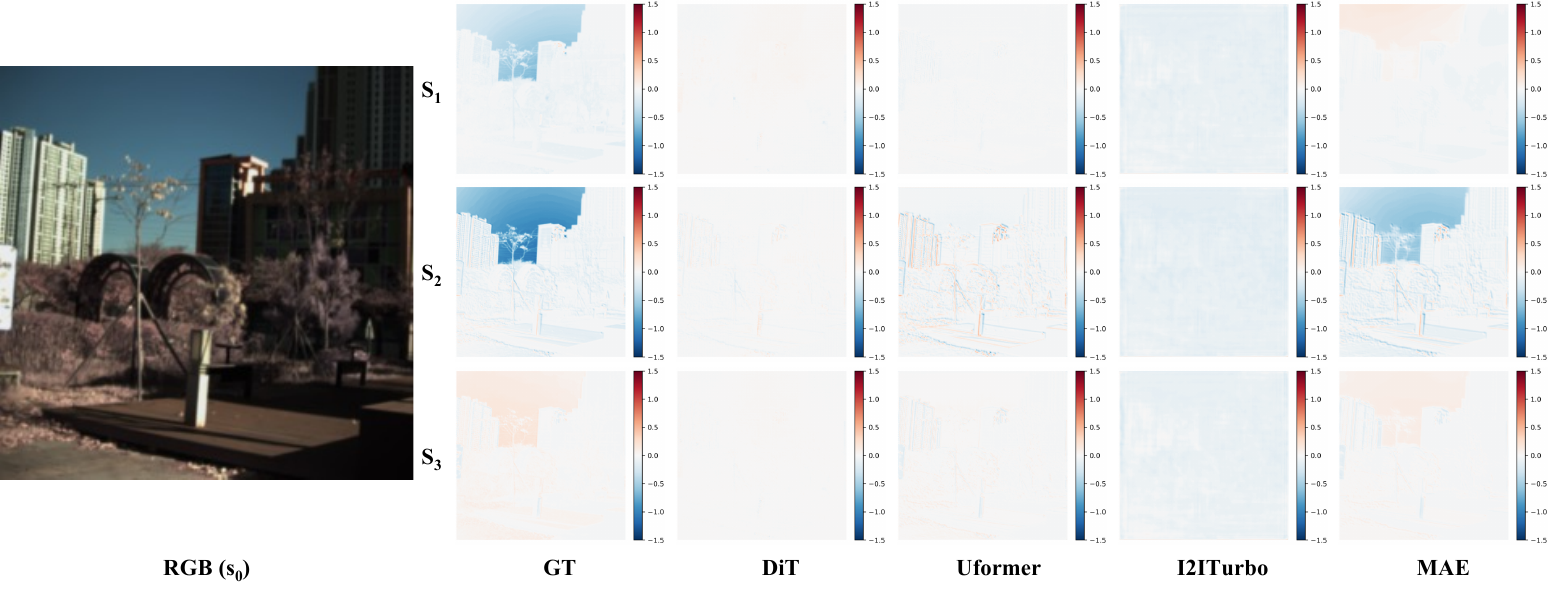}
    \caption{Reconstructed full Stokes components for test scene 1050, an urban plaza with high-rise buildings in which the sky region carries a strong negative $\mathrm{S}_2$ signal. MAE recovers $\mathrm{S}_2$ and $\mathrm{S}_3$ in this region but predicts the wrong sign for $\mathrm{S}_1$, while the other models predict nearly flat maps.}
    \label{fig_exp2}
\end{figure*}

\begin{figure*}[h]
    \centering
    \includegraphics[width=1\linewidth]{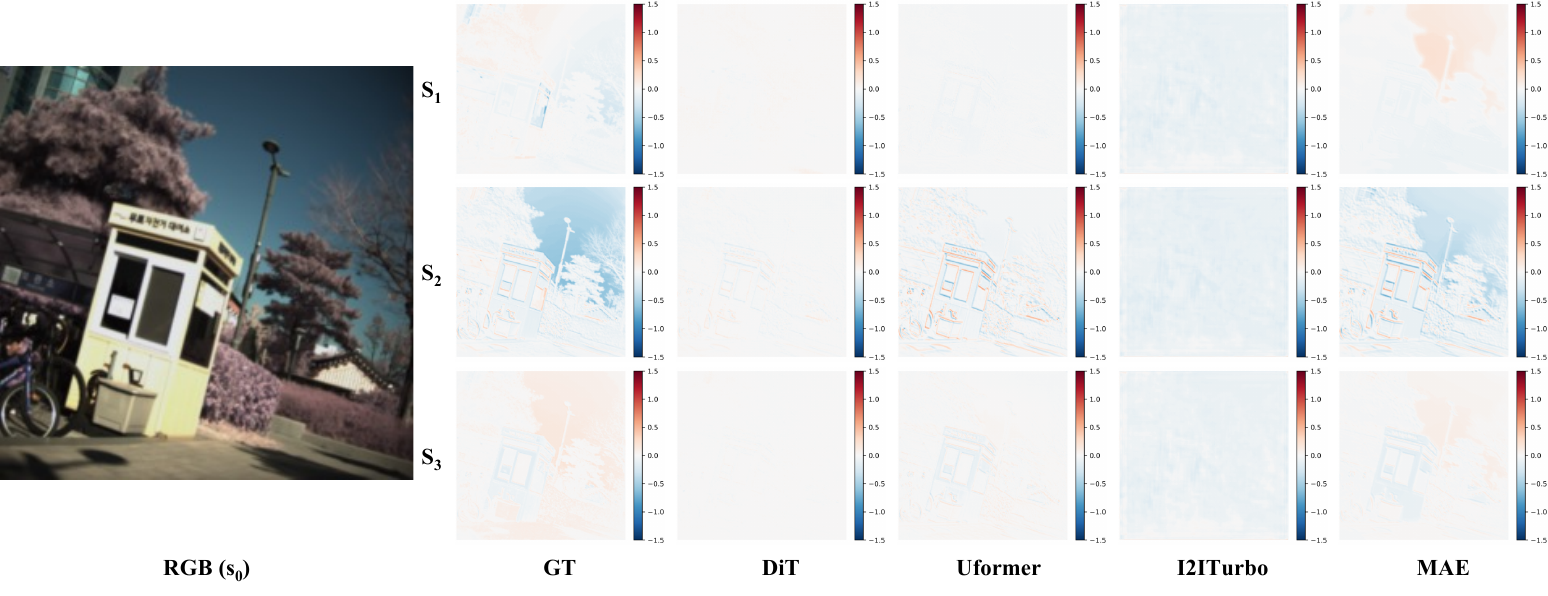}
    \caption{Reconstructed full Stokes components for test scene 1003.}
    \label{fig_exp3}
\end{figure*}

\begin{figure*}[h]
    \centering
    \includegraphics[width=1\linewidth]{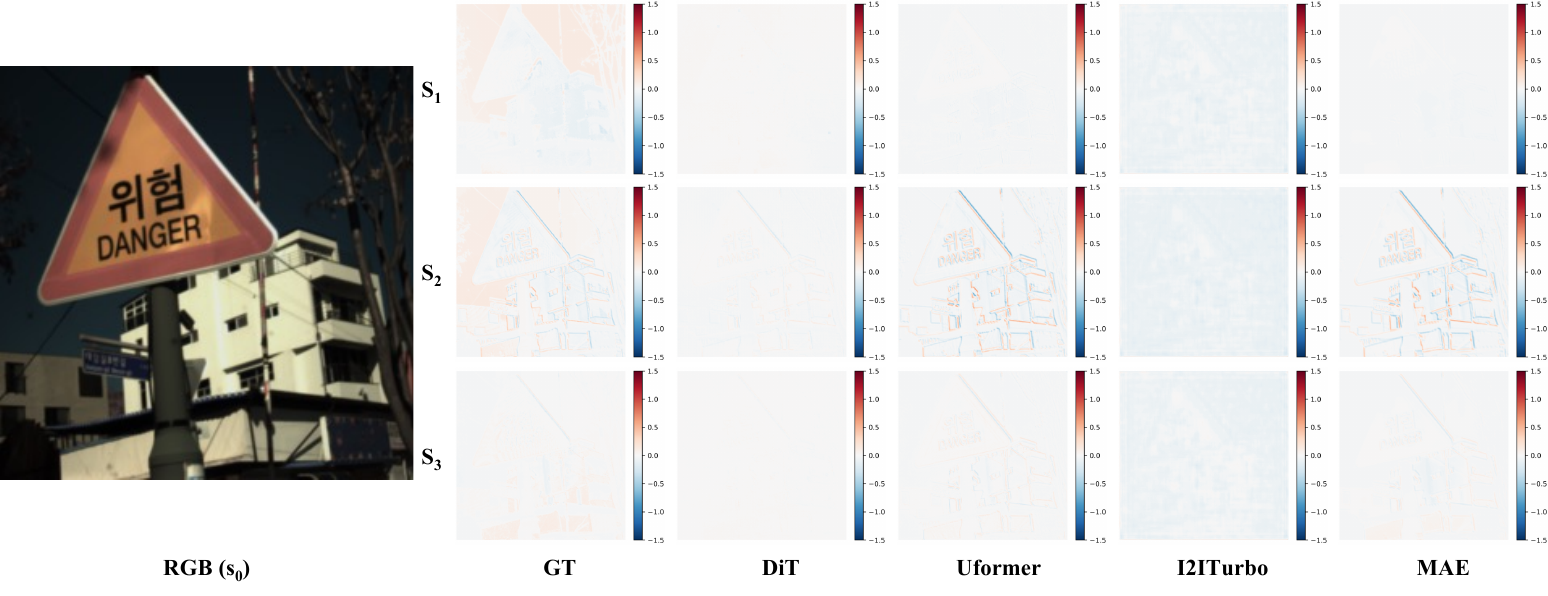}
    \caption{Reconstructed full Stokes components for test scene 1035.}
    \label{fig_exp4}
\end{figure*}

\begin{figure*}[t!]
    \centering
    \includegraphics[width=1\linewidth]{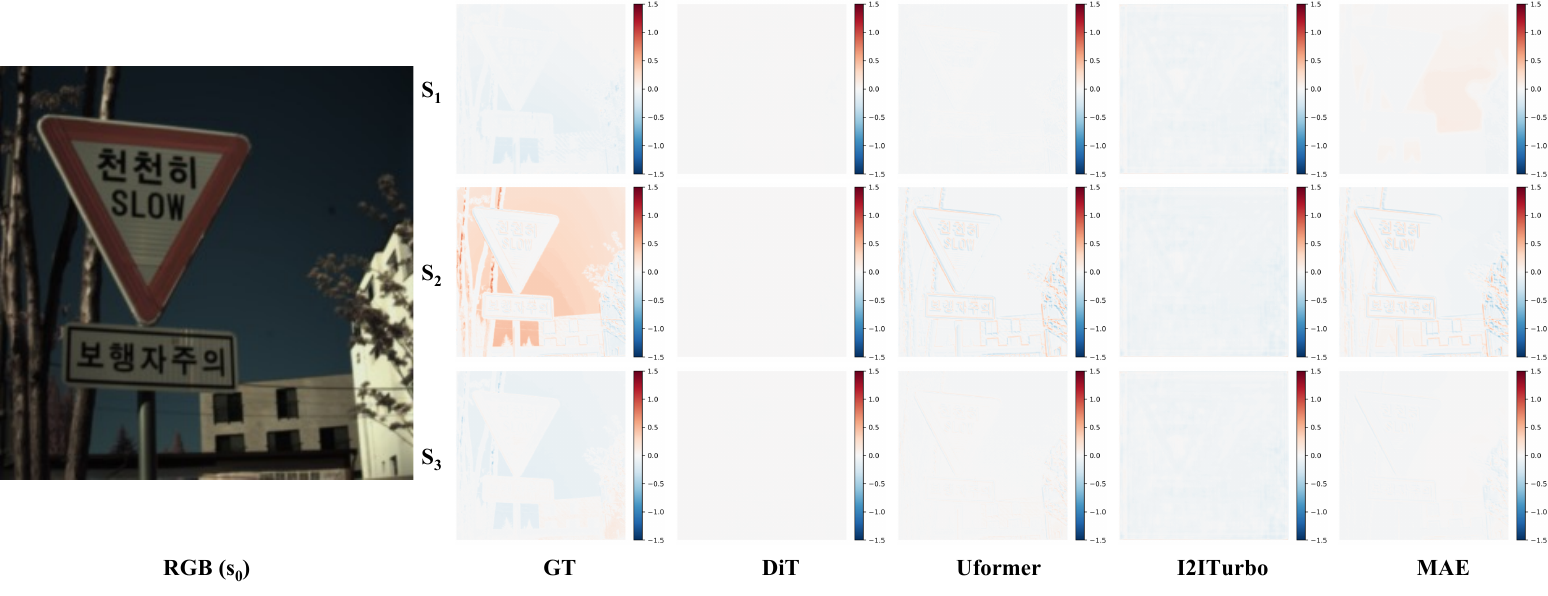}
    \caption{Reconstructed full Stokes components for test scene 1038.}
    \label{fig_exp5}
\end{figure*}

\begin{figure*}[t!]
    \centering
    \includegraphics[width=1\linewidth]{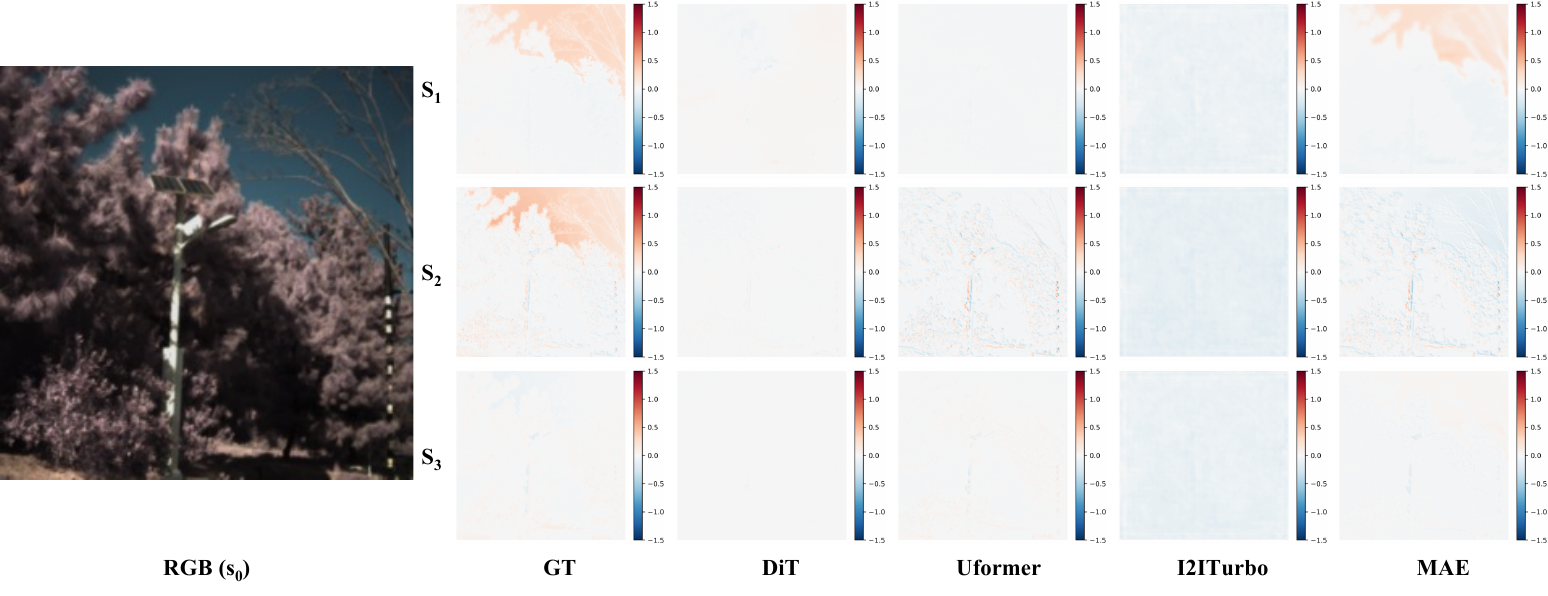}
    \caption{Reconstructed full Stokes components for test scene 1062.}
    \label{fig_exp6}
\end{figure*}

\begin{figure*}[t!]
    \centering
    \includegraphics[width=1\linewidth]{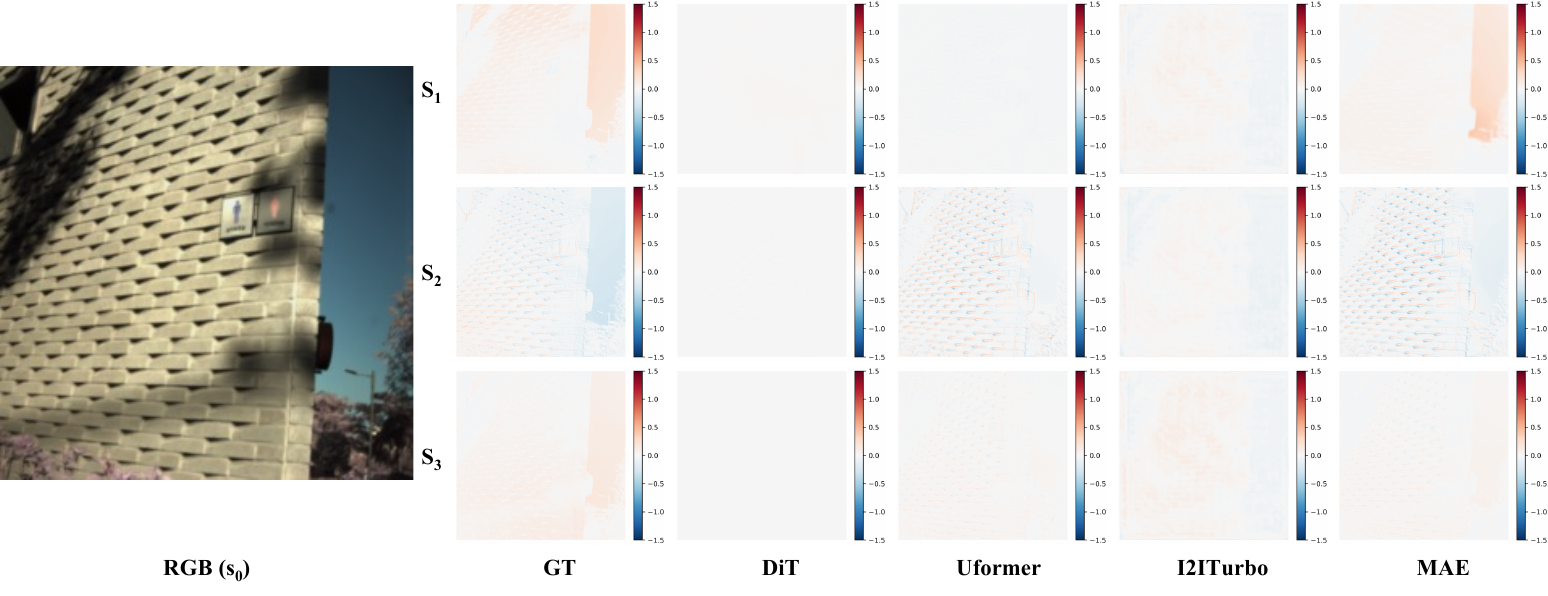}
    \caption{Reconstructed full Stokes components for test scene 1083.}
    \label{fig_exp7}
\end{figure*}

\begin{figure*}[t!]
    \centering
    \includegraphics[width=1\linewidth]{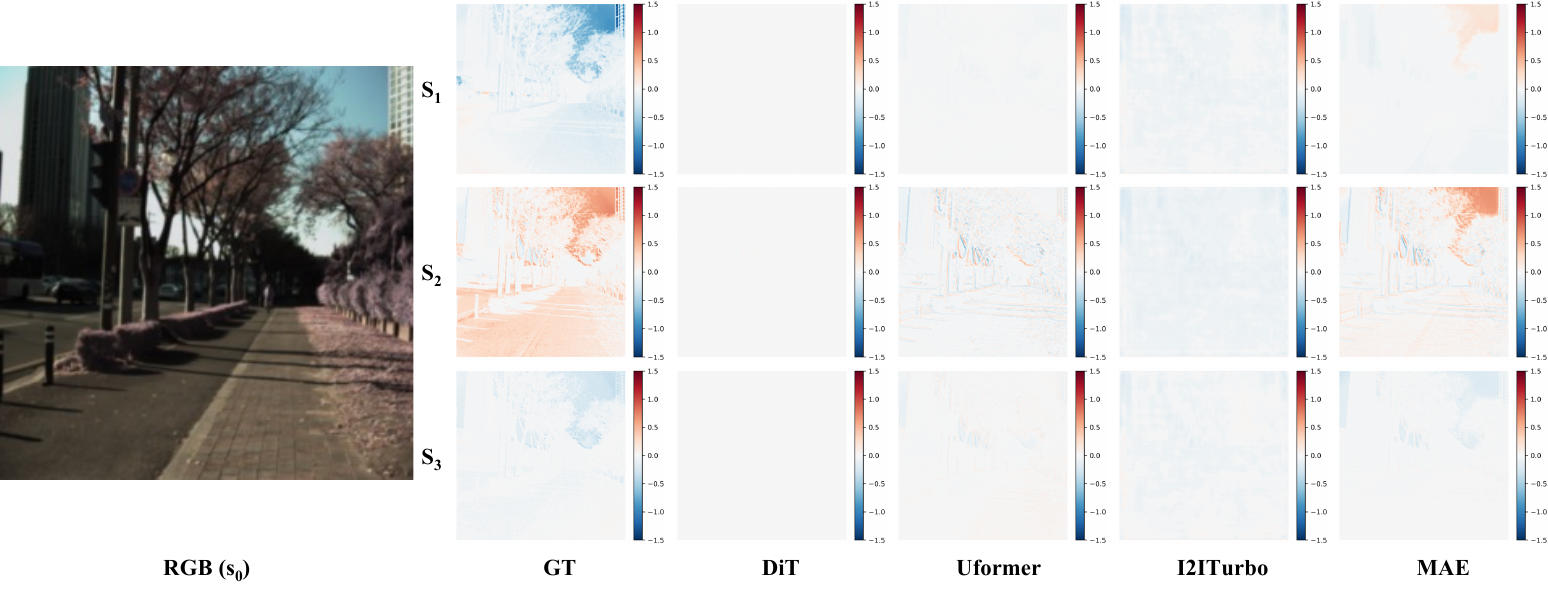}
    \caption{Reconstructed full Stokes components for test scene 1126.}
    \label{fig_exp8}
\end{figure*}

\begin{figure*}[t!]
    \centering
    \includegraphics[width=1\linewidth]{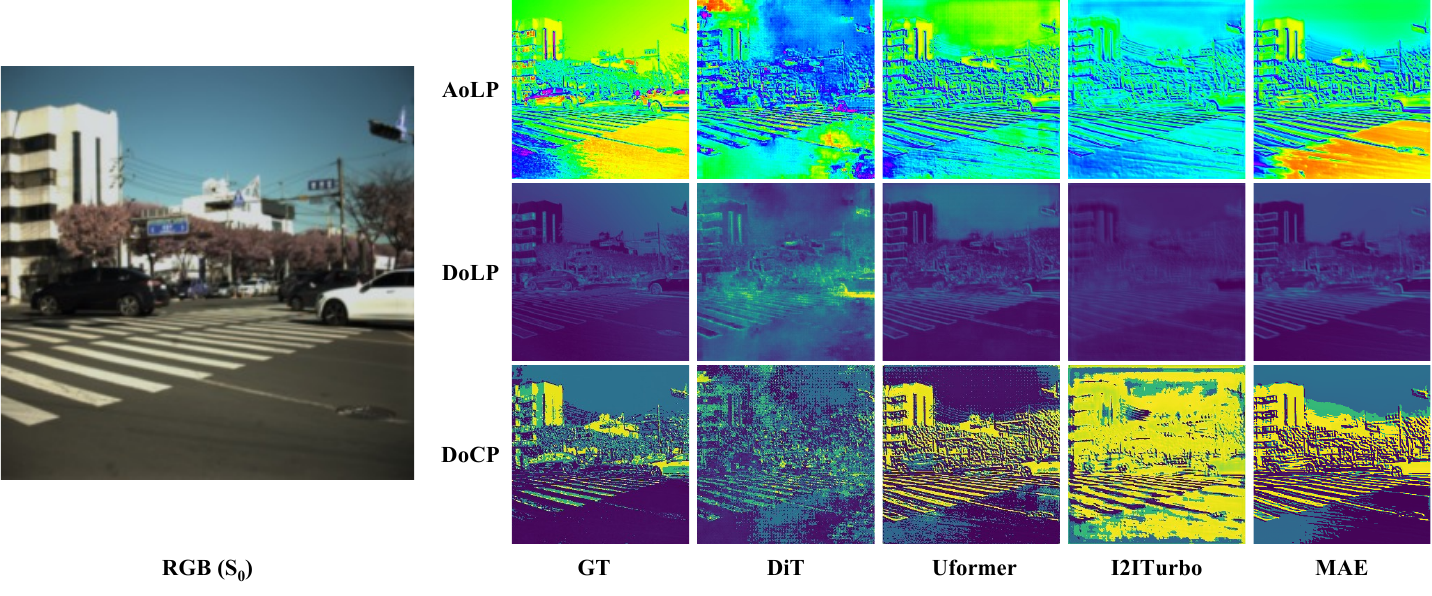}
    \caption{Predicted AoLP, DoLP, and DoCP for test scene 1158, a street intersection with crosswalk stripes, vehicles, and building facades. The models capture the global layout but miss abrupt polarization transitions; DiT produces noisy AoLP and I2ITurbo saturates DoCP, while MAE is closest to the ground truth.}
    \label{fig_dolp_aolp_001}
\end{figure*}

\begin{figure*}[t!]
    \centering
    \includegraphics[width=1\linewidth]{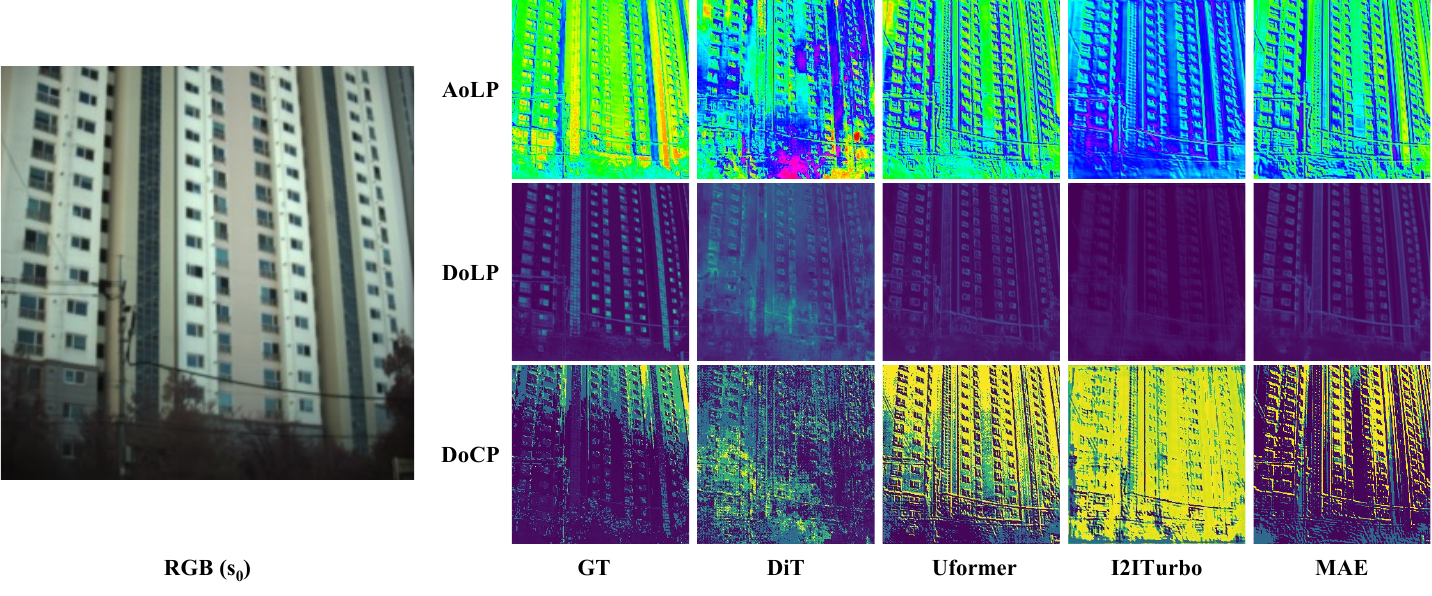}
    \caption{Predicted AoLP, DoLP, and DoCP for test scene 1196, a high-rise facade whose repetitive window pattern produces fine, periodic polarization structure.}
    \label{fig_dolp_aolp_002}
\end{figure*}

\begin{figure*}[t!]
    \centering
    \includegraphics[width=1\linewidth]{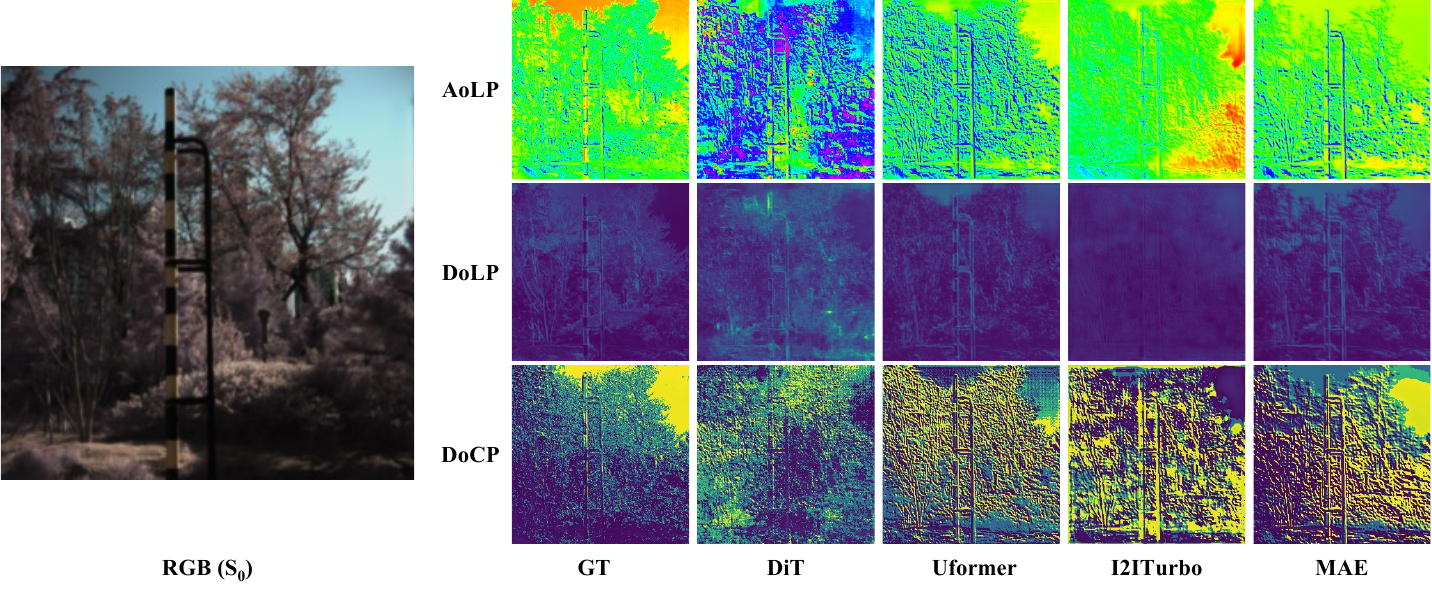}
    \caption{Predicted AoLP, DoLP, and DoCP for test scene 1036.}
    \label{fig_dolp_aolp_003}
\end{figure*}

\begin{figure*}[t!]
    \centering
    \includegraphics[width=1\linewidth]{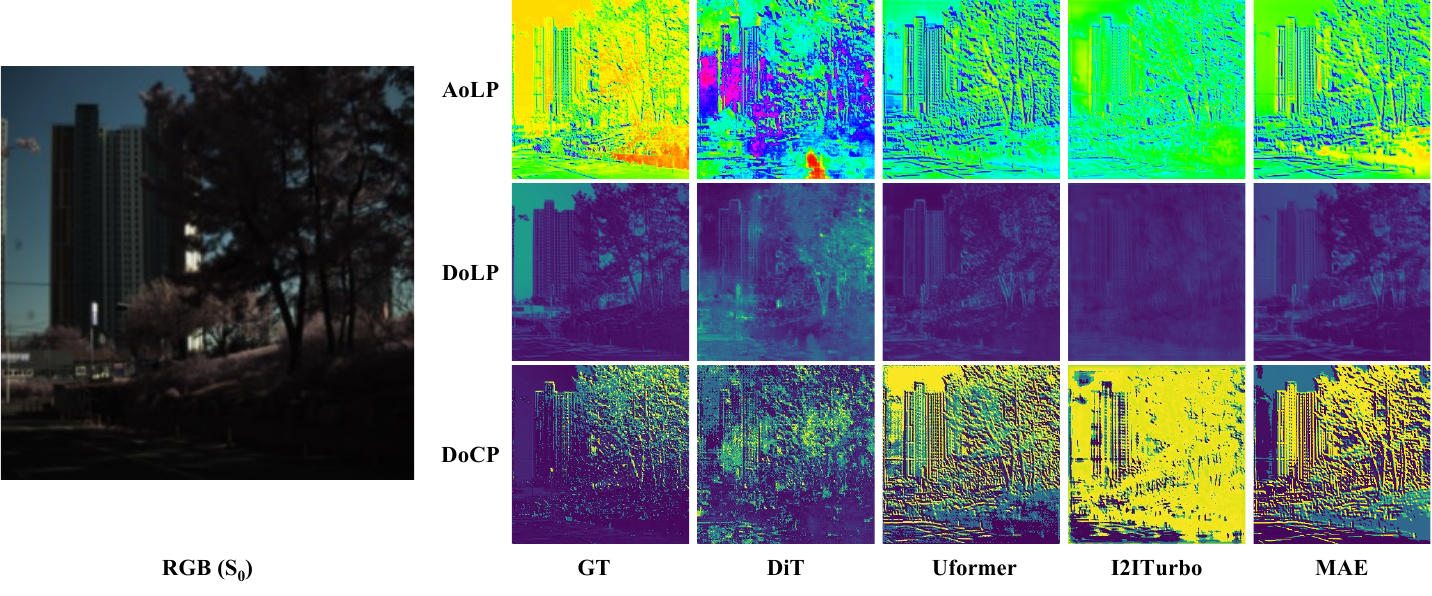}
    \caption{Predicted AoLP, DoLP, and DoCP for test scene 1040.}
    \label{fig_dolp_aolp_004}
\end{figure*}

\begin{figure*}[t!]
    \centering
    \includegraphics[width=1\linewidth]{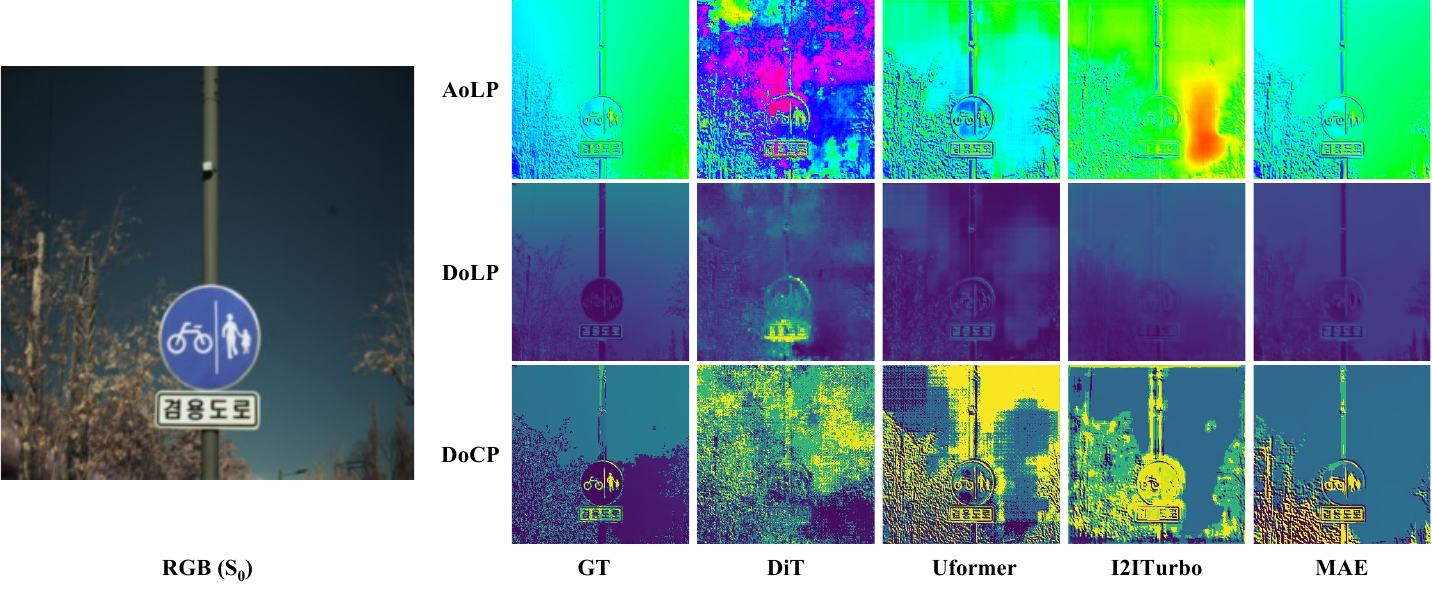}
    \caption{Predicted AoLP, DoLP, and DoCP for test scene 1047.}
    \label{fig_dolp_aolp_005}
\end{figure*}

\begin{figure*}[t!]
    \centering
    \includegraphics[width=1\linewidth]{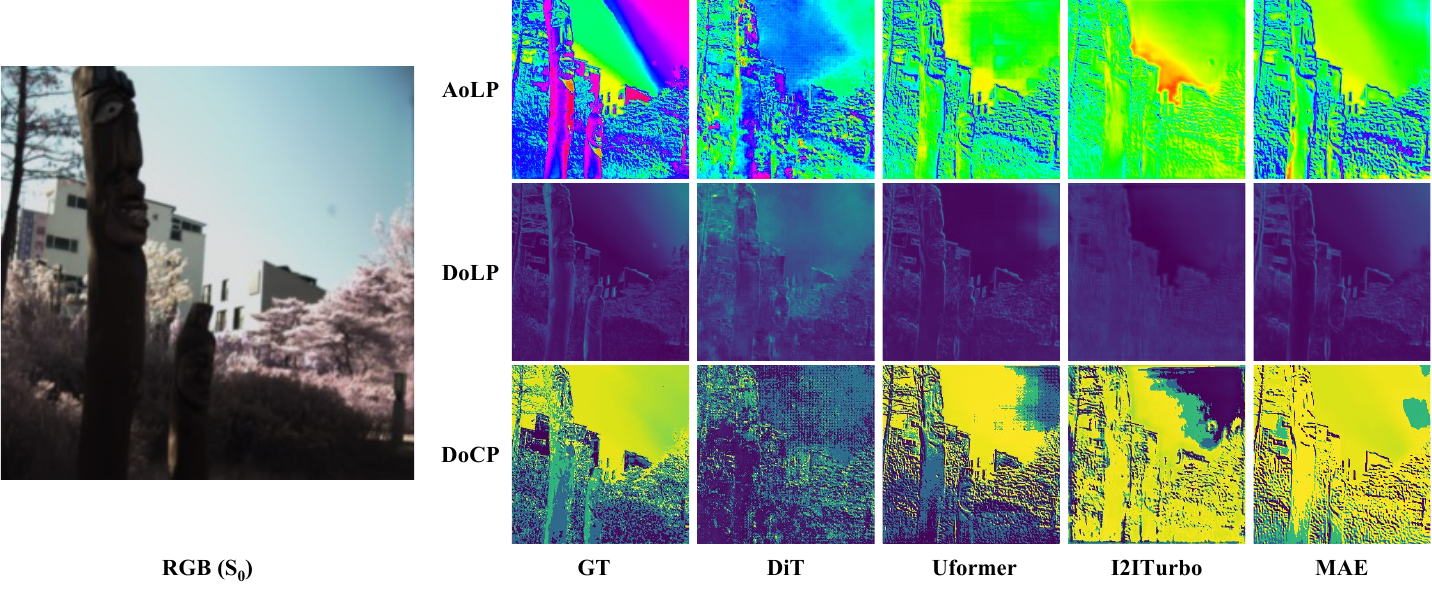}
    \caption{Predicted AoLP, DoLP, and DoCP for test scene 1052.}
    \label{fig_dolp_aolp_006}
\end{figure*}

\begin{figure*}[t!]
    \centering
    \includegraphics[width=1\linewidth]{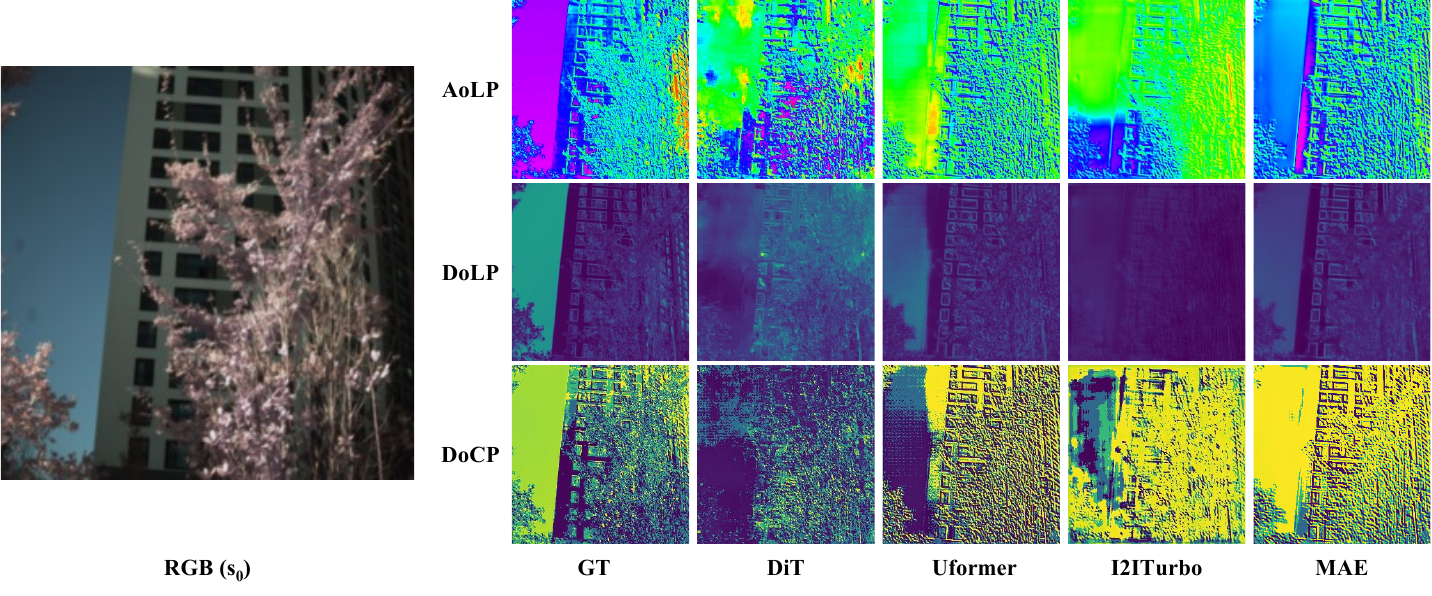}
    \caption{Predicted AoLP, DoLP, and DoCP for test scene 1194.}
    \label{fig_dolp_aolp_007}
\end{figure*}

Figures~\ref{fig_exp2}--\ref{fig_exp8} show reconstructed full Stokes components for further test scenes, and Figures~\ref{fig_dolp_aolp_001}--\ref{fig_dolp_aolp_007} show predicted AoLP, DoLP, and DoCP maps. In all figures, the input (left) is the normalized total-intensity image $\mathrm{s}_0$, and results are shown for DiT~\cite{peebles2023scalable}, Uformer~\cite{wang2022uformer}, I2ITurbo~\cite{parmar2024one}, and MAE~\cite{he2022masked}.

\section{Ethics statement and computational resources}
\label{app_ethics}

\paragraph{\textbf{Ethics statement}}
The work does not involve human subjects, and the data contain no annotations of people or other sensitive personal data. Outdoor scenes may incidentally show vehicles or bystanders, as in the source dataset~\cite{jeon2024spectral}. All data are used solely for research purposes.

\paragraph{\textbf{Computational resources}}
All experiments are performed on a server with eight NVIDIA RTX A5000 GPUs (24 GB memory each); training takes about one day per restoration model and two days per diffusion-based model.
The code, pre-trained checkpoints, configuration files, and the benchmark package are publicly available at \url{https://anonymous.4open.science/r/PolarScale}.

\end{document}